\documentclass[final,5p,times,twocolumn,authoryear]{elsarticle}

\usepackage{amssymb}
\usepackage{amsmath}

\usepackage{booktabs}

\usepackage{tabularx}
\usepackage{lettrine}
\usepackage{graphicx}
\usepackage{subcaption}

\journal{Information Processing \& Management}

\begin{document}

\begin{frontmatter}

\title{Cross-Modal Purification and Fusion for Small-Object RGB-D Transmission-Line Defect Detection\tnoteref{fund}}

\author[aff1]{Jiaming Cui}
\ead{hitcjm@stu.hit.edu.cn}

\author[aff2]{Wenqiang Li}
\ead{wenqli@stu.hit.edu.cn}

\author[aff2,aff3]{Shuai Zhou}
\ead{22B901046@stu.hit.edu.cn}

\author[aff2]{Ruifeng Qin}
\ead{qrf_951208@126.com}

\author[aff2]{Feng Shen\corref{cor1}}
\ead{fshen@hit.edu.cn}

\cortext[cor1]{Corresponding author.}

\affiliation[aff1]{organization={School of Mechatronics Engineering, Harbin Institute of Technology},
                  city={Harbin}, postcode={150001}, country={China}}
                  
\affiliation[aff2]{organization={School of Instrument Science and Engineering, Harbin Institute of Technology},
	city={Harbin}, postcode={150001}, country={China}}
                  
\affiliation[aff3]{organization={Electric Power Research Institute, Yunnan Power Grid Co., Ltd.},
                  city={Kunming}, postcode={650217}, country={China}}

\tnotetext[fund]{This work was supported by the National Natural Science Foundation of China under Grant 52371351, Grant 61673128, Grant 61573117, and Grant 41627801.}

\begin{abstract}
Automated defect detection in transmission lines is critical for power grid reliability but remains challenging because small-scale targets dominate UAV-acquired imagery. Most methods depend on RGB inputs alone, treating photometric appearance as the primary cue, a strategy that fails when defects show low chromatic contrast, ambiguous damage geometry, or partial occlusion by vegetation. Depth imagery offers complementary geometric evidence, yet fusing it with RGB is non-trivial: the two modalities differ markedly in noise characteristics and distributional properties, so na\"{i}ve combination can propagate modality-specific artifacts into the joint representation. We propose CMAFNet, a Cross-Modal Alignment and Fusion Network built on a purify-then-fuse paradigm for RGB-D defect detection. A Semantic Recomposition Module within each branch projects features through a learned bottleneck with position-wise normalization, suppressing sensor-specific noise while narrowing the inter-modal distribution gap. At the deepest fusion stage, a Contextual Semantic Integration Framework models global spatial dependencies via partial-channel attention, enabling the network to use structural priors such as regular insulator-string arrangement to distinguish small defects from similar background elements without the detail loss of full-channel attention. On the TL-RGBD benchmark, where 94.5\% of annotated instances are small objects, CMAFNet surpasses state-of-the-art detectors on all metrics. The full-scale model achieves 32.2\% mAP$_{50}$ and 12.5\% small-object AP, exceeding the best baseline by 9.8 and 4.0 points; the lightweight variant reaches 24.8\% mAP$_{50}$ at 228 FPS with 4.9M parameters. Ablation studies show the two modules interact synergistically with a combined gain of 13.7\% that exceeds their individual contributions, and confirm that depth provides geometric cues for boundary disambiguation and low-contrast localization that RGB alone cannot supply.
\end{abstract}




\end{frontmatter}

\section{Introduction}
\label{sec:introduction}

Transmission lines are among the most spatially distributed and environmentally exposed assets in modern power grids, continuously subjected to weathering, cyclic mechanical loading, and progressive material degradation such as corrosion and insulation aging \citep{zhangDSANetAttentionGuidedNetwork2024,luoUltrasmallBoltDefect2023}. Defects arising from these processes, including insulator contamination, conductor corrosion, and fitting damage, as well as external hazards such as bird nesting, can lead to flashovers, line tripping, and unplanned outages, increasing maintenance costs and compromising grid reliability \citep{baoDefectDetectionMethod2022,chengAdInDETRAdaptingDetection2024}. Timely identification of such defects is therefore critical for prioritizing maintenance scheduling, reducing unnecessary patrols, and mitigating safety risks \citep{yangDRANetDualBranchResidual2023a}. Because transmission networks often extend over thousands of kilometers, this task calls for automated visual inspection methods that can operate at high frequency and low marginal cost, requirements that are difficult to meet through manual or semi-automated workflows given limited workforce availability, access constraints in remote corridors, and the sheer volume of infrastructure to be covered.

Conventional inspection relies on human operators traversing transmission corridors on foot or by helicopter, or deploying robotic platforms such as line-crawling robots along predefined routes \citep{pouliotFieldorientedDevelopmentsLineScout2012,xuPowerLineGuidedAutomatic2022a}. These approaches are hazardous, labor-intensive, and costly to sustain at national-grid scale. The widespread adoption of unmanned aerial vehicles (UAVs) equipped with high-resolution cameras has substantially improved data acquisition, enabling efficient imagery collection over long corridors \citep{xuPowerLineGuidedAutomatic2022,dasilvaUnmannedAerialVehicle2020} and motivating a shift toward learning-based defect detection. The imagery obtained in this way, however, poses detection challenges that differ markedly from those in general-purpose benchmarks such as COCO or VisDrone. Transmission-line scenes are dominated by repetitive structural patterns and geometrically elongated components whose dense edge responses and strong background textures severely compress the effective pixel area available for defect discrimination. Defects typically occupy only a small fraction of the image, and this difficulty is compounded by oblique viewing angles, varying shooting distances, and persistent background clutter from vegetation, sky, and metallic infrastructure. On the TL-RGBD benchmark adopted in this study, approximately 94.5\% of annotated defect instances fall below the COCO small-object area threshold of $32 \times 32$ pixels at the $640 \times 640$ training input resolution, confirming that detection methods not explicitly designed for small objects tend to underperform in this domain.

Deep learning has driven substantial progress in transmission-line defect detection by exploiting the growing volume of UAV-acquired data. Existing approaches rely almost exclusively on RGB imagery as the sole input modality. Representative works span YOLO variants with attention-augmented backbones \citep{wangDeepLearningBased2024}, Faster R-CNN enhanced with aerial-image augmentation strategies \citep{caoAccurateGlassInsulators2023}, lightweight architectures with multiscale feature fusion \citep{maSmallSizedDefectDetection2025,wuCRLYOLOComprehensiveRecalibration2025}, and self-supervised vision transformer pipelines \citep{zhangTransmissionLineComponent2024}. A shared limitation of these methods is their dependence on photometric appearance, namely texture, color, and shading, as the primary discriminative cue. This dependence becomes a bottleneck when defects exhibit low chromatic contrast against cluttered backgrounds, when damage patterns are geometrically ambiguous, or when targets are partially occluded by vegetation \citep{heWeaklySupervisedContrastive2025,chenNovelIdentificationModel2025}. Because RGB images encode geometry only implicitly through shading and perspective rather than providing explicit three-dimensional surface cues, separating subtle structural deformations from benign surface variations remains difficult under such conditions \citep{liCompositeInsulatorOverheating2025,guptaLearningRichFeatures2014}. A further accuracy-efficiency tension constrains practical deployment: high-capacity models impose computational costs that exceed the processing budget of typical UAV-mounted hardware at the frame rates required for real-time inspection, while lightweight alternatives tend to sacrifice detection sensitivity on the small targets that dominate real-world transmission-line imagery \citep{tanEfficientDetScalableEfficient2020b,chenPRDeformableDETRDETR2024}.

Adding depth information to RGB appearance offers a promising path toward more discriminative representations. Depth maps encode relative surface geometry and spatial extent independently of illumination and texture, providing complementary evidence where photometric cues alone fall short, for instance by revealing surface discontinuities at damaged fittings or volumetric protrusions caused by bird nests. Fusing heterogeneous modalities, however, is not straightforward. RGB and depth features originate from fundamentally different sensing processes and exhibit distinct noise characteristics and distributional properties: depth maps may contain holes, quantization artifacts, and edge bleeding, while RGB images are susceptible to specular highlights and illumination variation. Direct concatenation or element-wise combination risks propagating these modality-specific artifacts into the joint representation, ultimately degrading rather than improving detection accuracy \citep{zhaoDeepMultimodalData2024}. How to align and fuse such heterogeneous representations, particularly under the combined pressures of modality-specific noise and the extreme small-object prevalence that characterizes transmission-line inspection, remains insufficiently addressed for this domain.

To address these challenges, this paper proposes CMAFNet, a Cross-Modal Alignment and Fusion Network that follows a purify-then-fuse paradigm for RGB-D feature integration. The core motivation is straightforward: fusing noisy, statistically misaligned representations tends to propagate their deficiencies rather than compensate for them. CMAFNet therefore introduces a dedicated purification stage before fusion. Within each modality branch, a Semantic Recomposition Module (SRM) reconstructs features through a structured bottleneck that suppresses modality-specific artifacts such as depth holes, edge bleeding, and RGB specular noise, while reducing distributional mismatch between the two streams. A Contextual Semantic Integration Framework (CSIF) then fuses these purified representations by capturing long-range structural dependencies that convolutional operators can only model at disproportionate cost, which is particularly relevant for reasoning about spatial relationships among small defects in densely repetitive patterns. To mitigate the detail erosion commonly associated with global attention, CSIF adopts a partial-channel design that balances contextual modeling with the fine-grained spatial discrimination needed for small-object localization. The resulting architecture is instantiated as a scalable model family, from compact configurations for real-time UAV inference to full-scale variants for server-side post-processing.

The core contributions of this work are threefold:
\begin{itemize}
	\item We introduce the purify-then-fuse paradigm for RGB-D feature integration in small-object detection and instantiate it as CMAFNet, a scalable architecture family spanning lightweight to full-scale configurations. CMAFNet achieves consistent improvements across all model scales on TL-RGBD, and systematic ablations confirm that purification before fusion yields measurable gains over direct-fusion alternatives.
	
	\item We propose the Semantic Recomposition Module (SRM), which suppresses modality-specific noise such as depth quantization artifacts and RGB illumination variations through a structured reconstruction bottleneck. Applied within each modality branch and at fused feature levels, SRM brings heterogeneous RGB and depth representations into more compatible statistical distributions, as validated by modality ablations and noise-sensitivity analyses in Section~\ref{sec:experiments}.
	
	\item We design the Contextual Semantic Integration Framework (CSIF), a partial-channel global attention mechanism that captures long-range structural dependencies while preserving spatial detail for small-object detection. By restricting global attention to a channel subset and retaining the remainder for local discrimination, CSIF addresses the tension between contextual coverage and spatial precision that limits conventional full-channel attention, as confirmed by component-level ablation.
\end{itemize}

The remainder of this paper is organized as follows. Section~\ref{sec:method} details the proposed CMAFNet architecture, including the dual-branch backbone, hierarchical fusion strategy, and the design of SRM and CSIF. Section~\ref{sec:experiments} presents the experimental setup, quantitative comparisons, and ablation analyses. Section~\ref{sec:conclusion} discusses limitations and future directions.

\section{Method}
\label{sec:method}

\subsection{Overview of Network Architecture}\label{sec:overview-of-network-architecture}

Combining depth geometry with RGB appearance can strengthen defect detection, but the gains depend heavily on how the two modalities interact during feature extraction and fusion. Naive strategies, whether early concatenation or late decision fusion, offer limited improvement because RGB and depth signals differ substantially in their statistical properties and noise characteristics \citep{zhaoDeepMultimodalData2024}. Early concatenation forces the network to reconcile heterogeneous distributions from the first layer onward, while late fusion defers cross-modal interaction to the prediction stage, leaving little room for the modalities to inform each other. A more effective strategy is to fuse modality-specific features at intermediate levels, where semantic abstraction is sufficient to tolerate noise yet spatial resolution still supports small-object localization. Prior methods that operate at this level, however, typically rely on static operations such as element-wise summation or channel concatenation followed by convolution. These fixed schemes cannot accommodate the dynamic variations common in aerial inspection: when RGB features are corrupted by illumination artifacts or depth features contain boundary discontinuities and quantization noise, static fusion propagates these corruptions into the shared representation, degrading detection performance rather than improving it.

\begin{figure*}[h]
	\centering
	\includegraphics[width=0.75\textwidth]{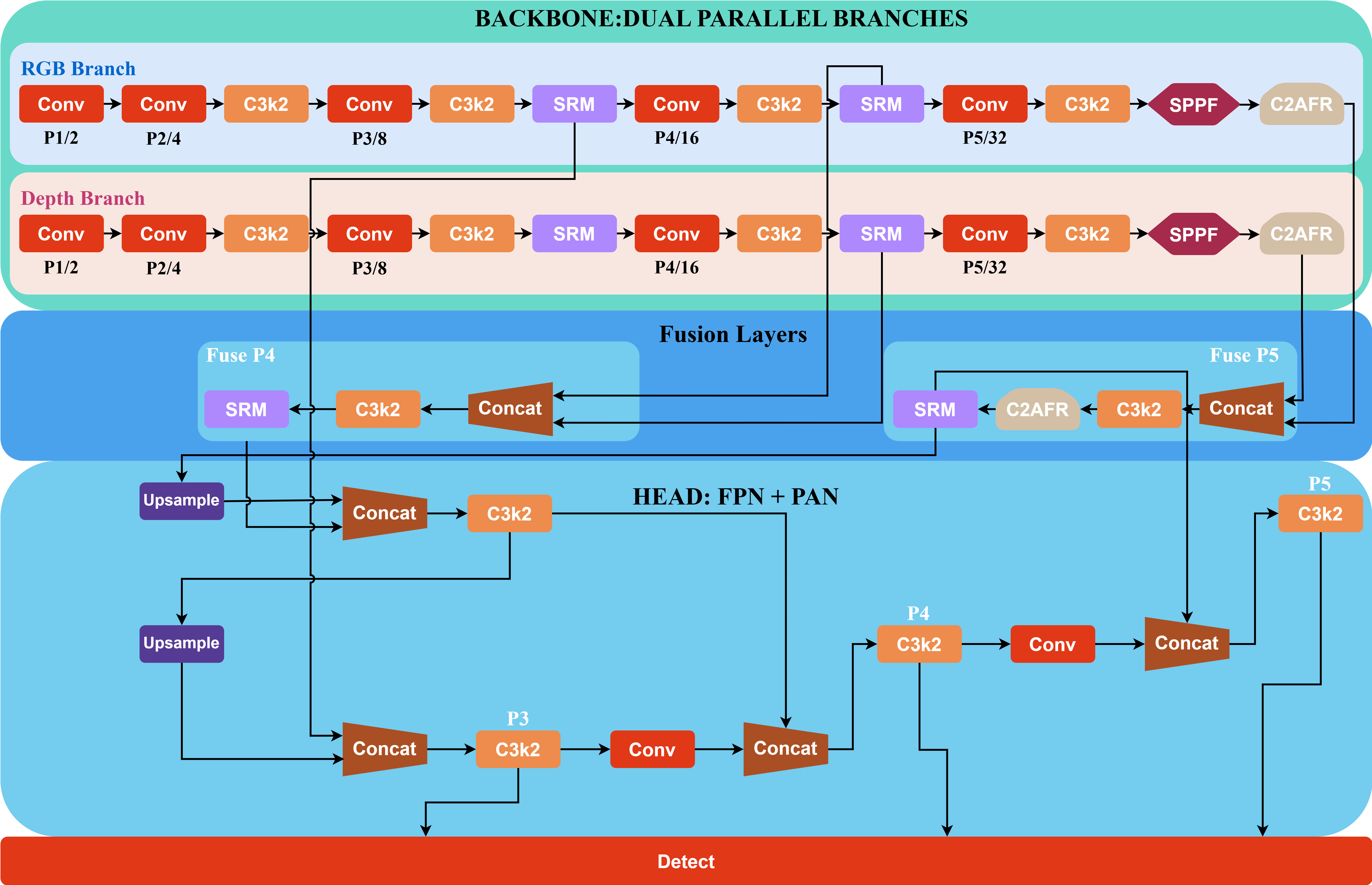}
	\caption{Overall architecture of CMAFNet. The dual-branch encoder processes RGB and depth modalities through symmetric but parameter-independent pathways, with SRM deployed at $P_3$ and $P_4$ for intra-modal feature purification. Cross-modal fusion occurs at $P_4$ and $P_5$; CSIF provides global attention-based context modeling at the $P_5$ fusion stage. The detection head implements bidirectional FPN-PAN feature propagation for multi-scale prediction. Depth features at $P_3$ are excluded from the detection pathway (dashed termination), so the highest-resolution prediction scale relies solely on RGB spatial detail.}
	\label{fig:overall_architecture}
\end{figure*}

These limitations call for a fusion design in which each modality stream first undergoes explicit noise suppression and distribution normalization before cross-modal integration, and in which the fused representation is then refined through context-aware mechanisms that reach beyond the local receptive field of standard convolutions. The proposed Cross-Modal Alignment and Fusion Network (CMAFNet) realizes this idea by extending the YOLO 11 \citep{jocherUltralyticsYOLO2023} single-stream detector into a dual-branch architecture for RGB-D input, while keeping all standard components, including C3k2, SPPF, the FPN-PAN neck, and the anchor-free decoupled detection head, unchanged. Three elements distinguish CMAFNet from the baseline: a Semantic Recomposition Module (SRM) that performs bottleneck-based feature purification with position-wise distribution normalization, a Contextual Semantic Integration Framework (CSIF) that introduces attention-driven global context modeling at the deepest fusion level, and a selective fusion strategy that restricts cross-modal integration to $P_4$ and $P_5$ while routing $P_3$ features exclusively from the RGB branch. Fig.~\ref{fig:overall_architecture} shows the complete network topology.

The data flow proceeds in three stages. In the first, the dual-branch backbone encodes RGB and depth inputs independently through to $P_5$, with SRM inserted at $P_3$ and $P_4$ within each branch to suppress modality-specific noise and narrow the inter-modal distribution gap. In the second, cross-modal fusion at $P_4$ and $P_5$ combines the purified branch features through channel concatenation, C3k2 re-encoding, and a second SRM pass; at $P_5$, CSIF is additionally placed between C3k2 and the terminal SRM to capture global context. In the third, the detection head receives the fused $P_4$/$P_5$ representations together with the RGB-only $P_3$ features through the standard FPN-PAN bidirectional pyramid for multi-scale prediction.

\subsubsection{Backbone: Dual Parallel Branches}
\label{sec:backbone}

The backbone consists of two encoding branches with identical topology but entirely separate parameters, one processing RGB input and the other processing depth input. All convolutional layers, C3k2 blocks, and the terminal SPPF module follow the default YOLO 11 \citep{jocherUltralyticsYOLO2023} implementation without modification.

The RGB branch receives 3-channel images, while the depth branch receives single-channel depth maps. Depth maps in the TL-RGBD dataset are spatially registered with their RGB counterparts through the factory calibration of the onboard RGB-D sensor (Intel RealSense D435i), and no additional geometric alignment is applied during preprocessing. Missing depth values and edge artifacts inherent to the sensor are not explicitly inpainted; robustness to these corruptions is instead delegated to SRM purification and the selective fusion strategy described in Section~\ref{sec:fusion}.

For the nano-scale configuration (width\_multiple $= 0.25$, depth\_multiple $= 0.50$, input $640 \times 640$), the feature hierarchy spans five pyramid levels: $P_1$ ($320{\times}320$, $C{=}16$), $P_2$ ($160{\times}160$, $C{=}32$), $P_3$ ($80{\times}80$, $C{=}64$), $P_4$ ($40{\times}40$, $C{=}128$), and $P_5$ ($20{\times}20$, $C{=}256$), with strides of 2, 4, 8, 16, and 32, respectively. The two branches produce identical feature dimensions at every level; the only asymmetry is that $P_3$ output from the depth branch is not propagated to the detection head.

Within each branch, an SRM instance ($K{=}128$, $\alpha{=}0.8$; see Section~\ref{sec:srm}) is inserted immediately after the C3k2 block at both $P_3$ and $P_4$. These scales offer a good balance: semantic content is rich enough for effective refinement, yet spatial resolution remains fine enough to preserve small-defect detail. Applying SRM before fusion narrows the inter-modal distribution gap at the source, providing cleaner inputs for the subsequent concatenation-based integration.

The rationale for parallel branches, rather than early fusion, lies in the distinct noise profiles of the two modalities. RGB imagery exhibits illumination variation and specular reflections, whereas depth maps suffer from edge bleeding, missing values at reflective surfaces, and quantization noise. Merging these signals before individual processing would entangle their noise distributions and complicate subsequent alignment. Independent encoding through $P_5$ allows each branch to develop modality-appropriate representations, deferring cross-modal integration to $P_4$ and $P_5$, where greater semantic abstraction makes noise-induced misalignment less problematic.

\subsubsection{Fusion Layers}
\label{sec:fusion}

The fusion architecture integrates RGB and depth features exclusively at $P_4$ and $P_5$. The $P_3$ output from the depth branch is disconnected from all downstream modules; the detection head receives $P_3$ input solely from the RGB branch after SRM processing. This is a fixed structural constraint rather than a learned gate. The resulting feature routing is as follows: the $P_3$ detection scale ($80{\times}80$) draws from the RGB branch through SRM (RGB only); the $P_4$ scale ($40{\times}40$) draws from \texttt{Concat}$\rightarrow$\texttt{C3k2}$\rightarrow$\texttt{SRM} (RGB + Depth); and the $P_5$ scale ($20{\times}20$) draws from \texttt{Concat}$\rightarrow$\texttt{C3k2}$\rightarrow$\texttt{CSIF}$\rightarrow$\texttt{SRM} (RGB + Depth).

Two observations motivate excluding depth from $P_3$. Depth features at this resolution carry the highest level of sensor noise and spatial misalignment; injecting them into the feature map most critical for small-defect localization would introduce more interference than useful geometric cues. Meanwhile, the primary contribution of depth, namely surface discontinuities and inter-component spatial layout, is adequately captured at $P_4$ and $P_5$, where feature granularity better matches the effective resolution of the sensor.

At $P_4$, the SRM-processed branch features ($C{=}128$ each) are concatenated along the channel dimension, yielding a 256-channel tensor. A C3k2 block re-encodes this tensor, and a subsequent SRM instance ($K{=}256$, $\alpha{=}0.8$) refines the fused representation, reinforcing cross-modally consistent patterns while attenuating modality-private residuals.

The $P_5$ pathway follows an analogous concatenation and C3k2 re-encoding procedure but additionally places CSIF ($n{=}2$ blocks; Section~\ref{sec:csif}) between C3k2 and the terminal SRM, forming the chain \texttt{Concat}$\rightarrow$\texttt{C3k2}$\rightarrow$\texttt{CSIF}$\rightarrow$\texttt{SRM}. At this depth, the two modalities often encode related semantic concepts with divergent statistics, and local convolutions alone cannot resolve the long-range dependencies needed for structural reasoning, for instance, recognizing the regular spacing of insulator strings to detect missing or damaged units. The terminal SRM applies a final round of distribution normalization to the globally refined features.

\subsubsection{Detection Head}

The detection head follows the standard YOLO 11 design without modification. In the FPN pathway, fused $P_5$ features are upsampled by a factor of $2\times$ (nearest-neighbor) and concatenated with fused $P_4$, then upsampled again and concatenated with RGB-only $P_3$; a C3k2 block refines the concatenated features at each level. The PAN pathway reverses this flow through stride-2 convolutions, propagating high-resolution spatial detail from $P_3$ back toward $P_4$ and $P_5$.

Each scale terminates in a decoupled head with separate classification and regression branches. Bounding boxes are predicted through distribution focal loss (DFL), which regresses discretized distributions over left-top-right-bottom offsets relative to anchor points. Positive samples are assigned by the Task-Aligned Assigner, which scores candidates jointly on classification confidence and localization quality. The training loss combines binary cross-entropy for classification, CIoU for box regression, and DFL for distribution regression, all with default YOLO 11 weights. The proposed contributions are thus confined entirely to the feature extraction and fusion stages; the detection head and training procedure remain identical to those of the baseline.

\subsection{Semantic Recomposition Module}
\label{sec:srm}

SRM is deployed at six locations: $P_3$ and $P_4$ within each branch (four instances, $K{=}128$) and the fused $P_4$ and $P_5$ levels (two instances, $K{=}256$). All instances share the same four-stage architecture---encode, refine, normalize, decode---followed by residual injection, differing only in the latent dimension $K$.

\begin{figure}[h]
	\centering
	\includegraphics[width=0.9\columnwidth]{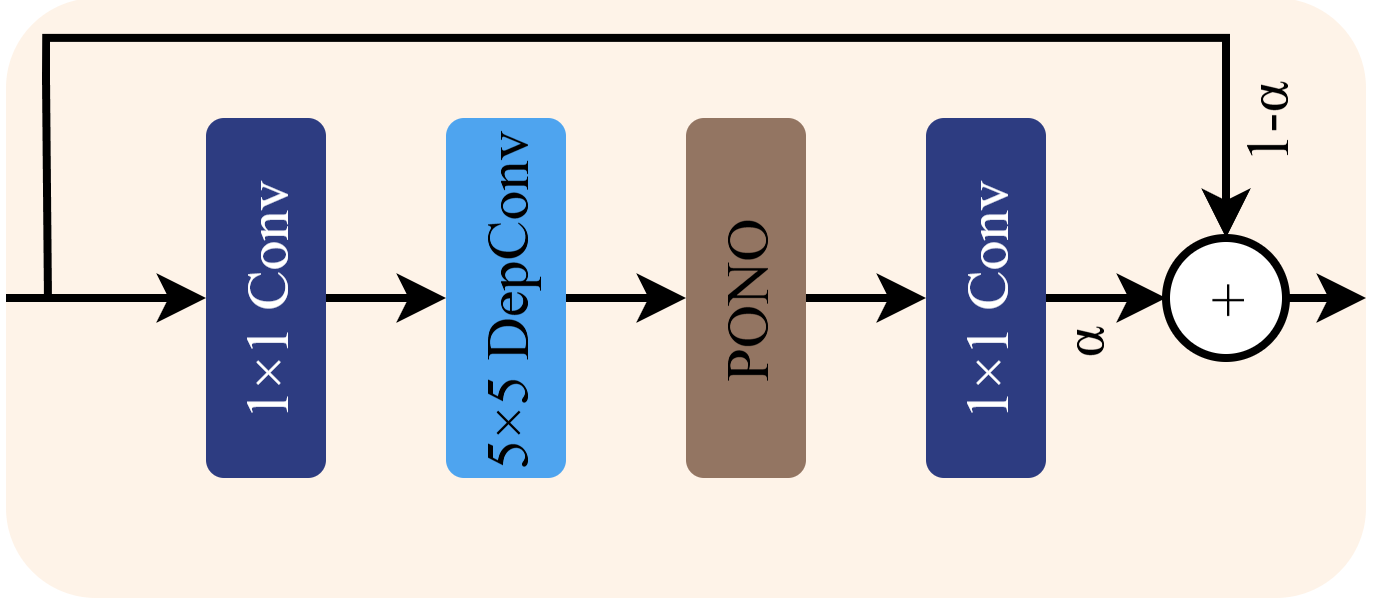}
	\caption{Architecture of the Semantic Recomposition Module. The four-stage pipeline consists of: (1) pointwise encoding from $C$ to $K$ channels, (2) $5 \times 5$ depthwise refinement without activation, (3) position-wise normalization (PONO) across the $K$ dimension, and (4) pointwise decoding from $K$ back to $C$ channels. The output is a convex combination of the refined and original features controlled by the mixing coefficient $\alpha$.}
	\label{fig:srm}
\end{figure}

SRM suppresses modality-specific noise without eroding the fine-grained details on which small-defect localization depends. In transmission-line imagery, background textures such as conductor strand patterns and fitting reflections, structural edges at conductor--sky boundaries, and depth sensor artifacts including holes, edge bleeding, and quantization noise all produce strong feature responses that inflate false-positive rates at the scales most relevant to small targets. Channel and spatial attention mechanisms can reweight existing activations but cannot impose structural constraints on what constitutes valid content. SRM takes a different approach: it compresses features into a learned low-dimensional subspace, refines them there, and reinjects the result through controlled residual mixing.

Given an input feature map $\mathbf{F} \in \mathbb{R}^{B \times C \times H \times W}$, the pipeline proceeds as follows (Fig.~\ref{fig:srm}). The encoding stage projects $\mathbf{F}$ into a $K$-dimensional latent space via a $1 \times 1$ pointwise convolution without batch normalization or activation:
\begin{equation}
	\mathbf{E} = \mathcal{P}_{\mathrm{enc}}(\mathbf{F}), \quad \mathcal{P}_{\mathrm{enc}}: \mathbb{R}^{C} \rightarrow \mathbb{R}^{K}, \quad K < C.
\end{equation}
The reduction from $C$ to $K$ creates an information bottleneck that retains only components with a stable low-dimensional representation while discarding incidental variation.

The refinement stage aggregates local spatial context within the latent space through a $5 \times 5$ depthwise convolution applied independently to each of the $K$ channels:
\begin{equation}
	\mathbf{R} = \mathcal{D}_{5 \times 5}(\mathbf{E}).
\end{equation}
No activation function follows this convolution; keeping the refinement stage linear prevents nonlinear coupling of noise across latent channels. The $5 \times 5$ kernel is large enough to capture local relationships among adjacent structural elements while remaining parameter-efficient thanks to the depthwise formulation.

The normalization stage applies position-wise standardization across the $K$ dimension at each spatial location:
\begin{equation}
	\mu_{h,w} = \frac{1}{K} \sum_{k=1}^{K} \mathbf{R}_{k,h,w}, \quad \sigma_{h,w} = \sqrt{\frac{1}{K} \sum_{k=1}^{K} \left( \mathbf{R}_{k,h,w} - \mu_{h,w} \right)^2 + \epsilon},
\end{equation}
\begin{equation}
	\hat{\mathbf{R}}_{k,h,w} = \frac{\mathbf{R}_{k,h,w} - \mu_{h,w}}{\sigma_{h,w}},
\end{equation}
where $\epsilon = 10^{-5}$. Unlike batch or layer normalization, which pool statistics over spatial or batch dimensions, position-wise normalization computes statistics at each $(h,w)$ location independently. Transmission-line images exhibit strong spatial heterogeneity: sky regions, metallic towers, and insulator assemblies produce markedly different activation patterns, and sharing statistics across these regions would suppress informative variation. A side effect of this design is that absolute activation magnitudes are lost at each position, leaving only the relative distribution across latent channels. This reduces sensitivity to the systematic scale gap between RGB and depth activations at corresponding locations, but also discards magnitude information that may matter for confidence estimation. The residual mixing in the final stage partly compensates for this loss by retaining a fraction of the original features.

The decoding stage maps the normalized representation back to $C$ channels via a $1 \times 1$ pointwise convolution without batch normalization or activation:
\begin{equation}
	\mathbf{D} = \mathcal{P}_{\mathrm{dec}}(\hat{\mathbf{R}}), \quad \mathcal{P}_{\mathrm{dec}}: \mathbb{R}^{K} \rightarrow \mathbb{R}^{C}.
\end{equation}
The output is formed as a convex combination of the refined and original features:
\begin{equation}
	\mathbf{Y} = \alpha \cdot \mathbf{D} + (1 - \alpha) \cdot \mathbf{F},
\end{equation}
where $\alpha = 0.8$ is a fixed scalar. Giving most of the weight to the refined branch maximizes noise suppression, while the $(1{-}\alpha)$ residual preserves fine-grained spatial detail that the bottleneck cannot fully reconstruct, balancing purification strength against localization fidelity.

All six instances share the same architecture but play distinct roles depending on $K$. At the branch level ($K{=}128$), the compact bottleneck filters modality-private noise, such as illumination artifacts in RGB and sensor holes together with quantization errors in depth, by compressing features before normalization. This yields per-modality representations with narrower inter-modal distribution gaps. At the fusion level ($K{=}256$), the wider bottleneck operates on concatenated cross-modal features; position-wise normalization then acts on the joint RGB-depth latent space, attenuating channels whose activations are inconsistent across modalities while reinforcing those that show coherent patterns. In both cases, $K$ is set to roughly half the input channel dimension, trading off compression strength against representational capacity.

\subsection{Contextual Semantic Integration Framework}
\label{sec:csif}

Because SRM relies on local convolutions ($5{\times}5$ depthwise kernel) and per-position normalization, it cannot capture long-range spatial dependencies. At the $P_5$ fusion level, however, defect discrimination often depends on global structural context: a damaged insulator is easier to identify when the network can reference the regular arrangement of intact neighbors, and a bird nest is distinguished from legitimate equipment by its inconsistency with the expected component layout. To provide this capability, CMAFNet inserts CSIF at the $P_5$ fusion stage, between C3k2 and the terminal SRM.

CSIF is placed exclusively at $P_5$ for two reasons. Global attention is most effective when features encode abstract relational semantics rather than fine-grained spatial patterns. In addition, the quadratic cost of self-attention ($\mathcal{O}(H^2 W^2)$) is tractable only at the lowest resolution ($20{\times}20$ at the nano scale); deploying CSIF at $P_3$ or $P_4$ would multiply FLOPs by $16\times$ or $4\times$, respectively, without a commensurate gain.

\begin{figure}[h]
	\centering
	\includegraphics[width=0.85\columnwidth]{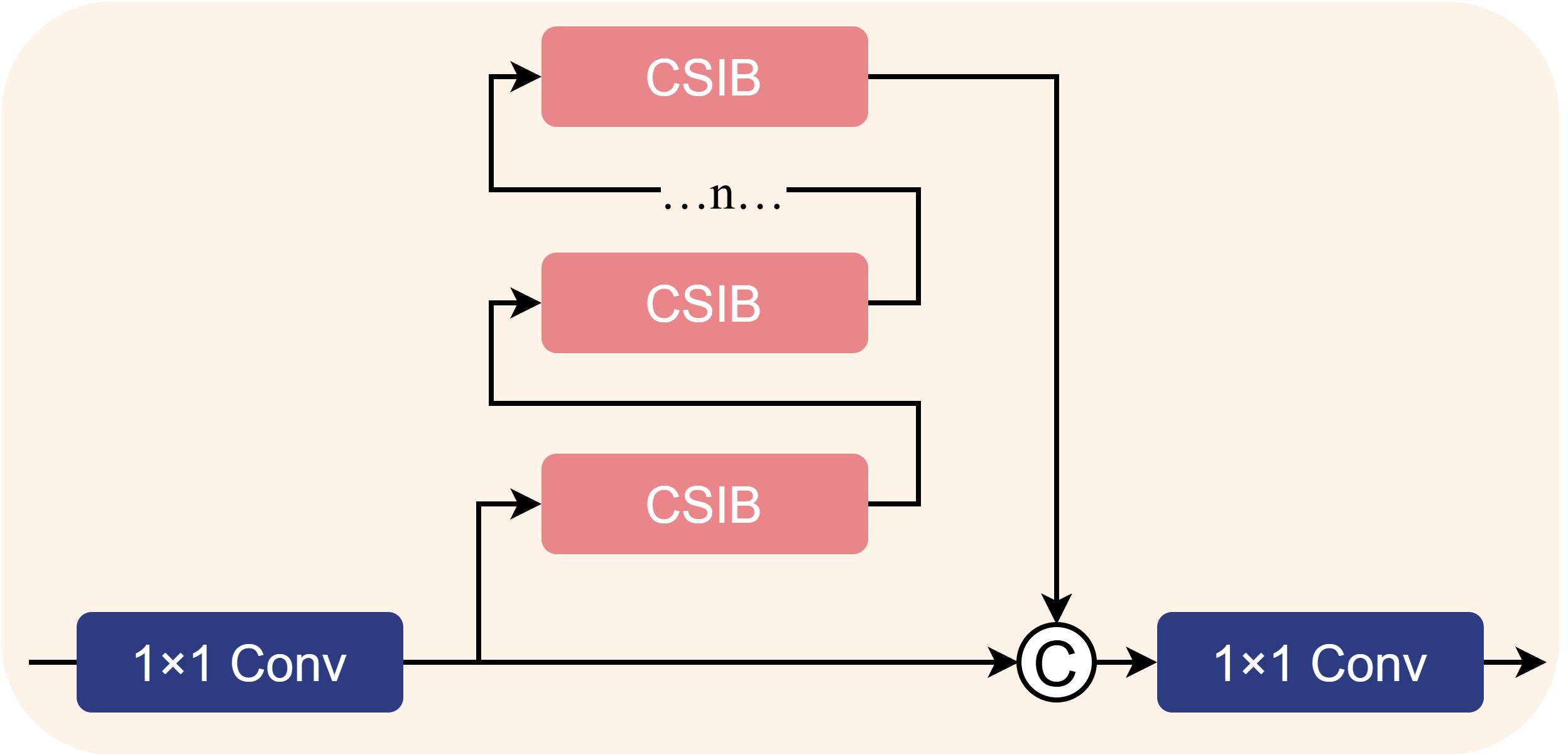}
	\caption{Architecture of the Contextual Semantic Integration Framework. A convolutional stem expands channel dimensions; the resulting feature map is split along the channel axis into an attention portion and a bypass portion. The attention portion is refined by $n$ stacked CSIBs, then concatenated with the bypass and projected to the output dimension.}
	\label{fig:csif}
\end{figure}

As illustrated in Fig.~\ref{fig:csif}, CSIF processes an input $\mathbf{X} \in \mathbb{R}^{B \times C_{\mathrm{in}} \times H \times W}$ through a project--refine--project pipeline. A $1{\times}1$ convolutional stem with BN and SiLU expands the channel dimension by a factor of $e{=}2$, producing a $2C_{\mathrm{in}}$-channel tensor that is then split into two equal halves along the channel axis. One half, the attention portion, passes through a sequence of $n$ Contextual Semantic Interaction Blocks (CSIBs), while the other, the bypass portion, is forwarded without modification. After $n$ iterations, the two halves are concatenated, and a $1{\times}1$ convolution with BN and SiLU maps the result to the target dimension $C_{\mathrm{out}}$. Restricting attention to half the channels confines global interaction to a subspace of the representation, which mitigates the over-smoothing that full-channel attention would impose on small-object features and halves the computational cost of the attention operation.

\begin{figure}[h]
	\centering
	\includegraphics[width=0.85\columnwidth]{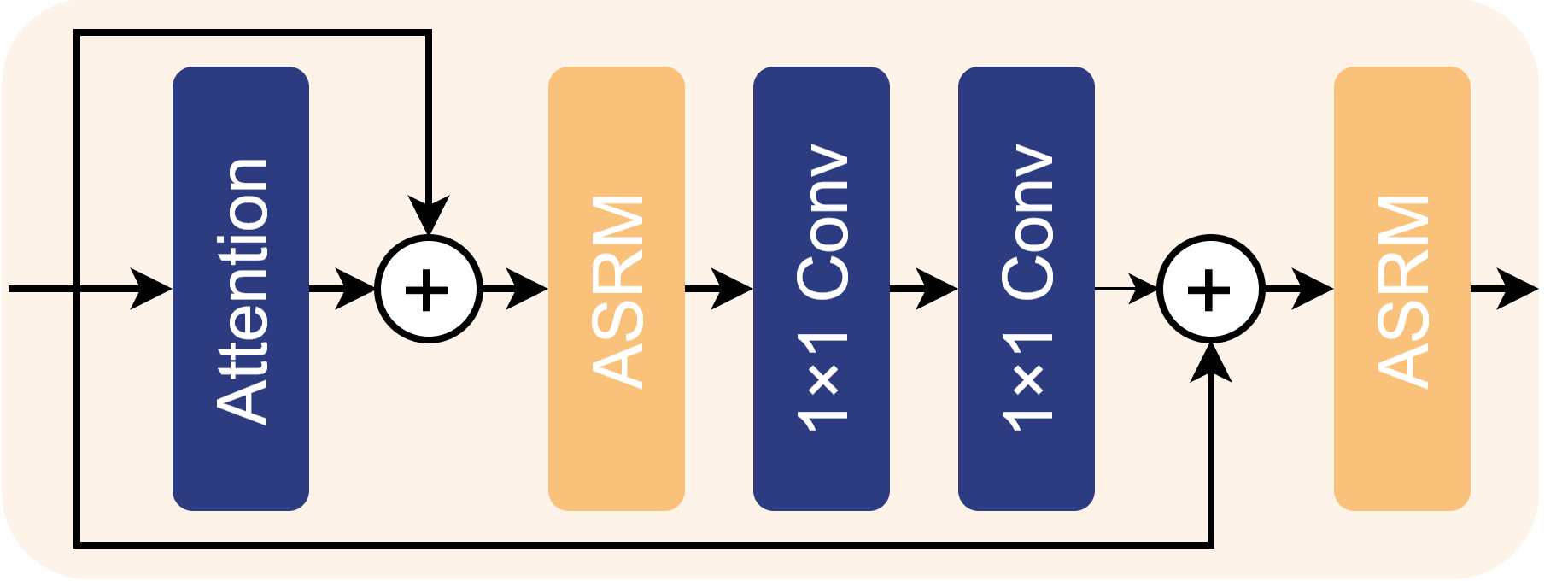}
	\caption{Structure of the Contextual Semantic Interaction Block. Each block applies multi-head self-attention followed by ASRM, then a feed-forward network followed by ASRM. Residual connections are present at both stages.}
	\label{fig:csib}
\end{figure}

Each CSIB (Fig.~\ref{fig:csib}) follows a standard attention--FFN cycle, with ASRM replacing conventional layer normalization in a post-normalization arrangement. The first phase computes multi-head self-attention:
\begin{equation}
	\mathbf{Q} = \mathbf{H} \mathbf{W}_Q, \quad \mathbf{K} = \mathbf{H} \mathbf{W}_K, \quad \mathbf{V} = \mathbf{H} \mathbf{W}_V,
\end{equation}
\begin{equation}
	\mathrm{Attn}(\mathbf{H}) = \mathrm{softmax}\!\left( \frac{\mathbf{Q} \mathbf{K}^\top}{\sqrt{d_k}} \right) \mathbf{V},
\end{equation}
with $h = \max(1, \lfloor C / 64 \rfloor)$ heads and a per-head dimension of $d_k = C / h$. A residual connection yields $\mathbf{U} = \mathbf{H} + \mathrm{Attn}(\mathbf{H})$, followed by ASRM: $\hat{\mathbf{U}} = \mathrm{ASRM}(\mathbf{U})$. The second phase applies a two-layer feed-forward network:
\begin{equation}
	\mathrm{FFN}(\hat{\mathbf{U}}) = \mathbf{W}_2 \, \sigma(\mathbf{W}_1 \hat{\mathbf{U}} + \mathbf{b}_1) + \mathbf{b}_2,
\end{equation}
where $\sigma$ denotes SiLU and the hidden dimension is $2C$. A second residual connection and ASRM produce the block output: $\mathbf{H}^{(i)} = \mathrm{ASRM}(\hat{\mathbf{U}} + \mathrm{FFN}(\hat{\mathbf{U}}))$. Post-normalization is preferred over pre-normalization because the concatenated RGB-depth features exhibit persistent distribution heterogeneity; normalizing after the residual addition corrects this distributional shift before the next transformation can amplify it.

ASRM regulates the distribution heterogeneity of cross-modal fused features more flexibly than standard layer normalization. Given $\mathbf{x} \in \mathbb{R}^{B \times C \times H \times W}$, it first computes a gated interpolation between the layer-normalized representation and the original input:
\begin{equation}
	\tilde{\mathbf{x}} = \boldsymbol{\gamma} \odot \mathrm{LN}(\mathbf{x}) + \boldsymbol{\gamma}_x \odot \mathbf{x},
	\label{eq:asrm_gate}
\end{equation}
where $\mathrm{LN}(\cdot)$ denotes channel-wise layer normalization with statistics computed over $C$ at each spatial position, $\boldsymbol{\gamma}, \boldsymbol{\gamma}_x \in \mathbb{R}^{C \times 1 \times 1}$ are learnable per-channel scales initialized to $\boldsymbol{\gamma} {=} 0.01$ and $\boldsymbol{\gamma}_x {=} 1$, and $\odot$ denotes the Hadamard product. Under this initialization, ASRM approximates an identity mapping at the start of training, avoiding premature interference with the still-unstable attention weights. As training proceeds, the gate gradually shifts toward the normalized branch, producing a smooth transition from weak to strong regulation.

A constrained adaptation mapping $\mathcal{F}$ then computes a low-rank residual correction on the gated features:
\begin{equation}
	\mathcal{F}(\tilde{\mathbf{x}}) = \mathcal{P}_{\mathrm{up}}\!\Big(\mathrm{Dropout}\!\big(\sigma\!\big(\mathrm{DW}_{3,5,7}\!\big(\mathcal{P}_{\mathrm{down}}(\tilde{\mathbf{x}})\big)\big)\big)\Big),
\end{equation}
where $\mathcal{P}_{\mathrm{down}}{:}\;\mathbb{R}^{C}{\to}\mathbb{R}^{r}$ is a $1{\times}1$ convolution that reduces channels to $r{=}64$, $\mathrm{DW}_{3,5,7}$ denotes three parallel depthwise convolutions with kernel sizes of $3{\times}3$, $5{\times}5$, and $7{\times}7$ whose outputs are summed, $\sigma$ is GELU, Dropout operates with a probability of 0.05, and $\mathcal{P}_{\mathrm{up}}{:}\;\mathbb{R}^{r}{\to}\mathbb{R}^{C}$ restores the original channel dimension. The final output is:
\begin{equation}
	\mathbf{y} = \mathbf{x} + \mathcal{F}(\tilde{\mathbf{x}}).
	\label{eq:asrm_residual}
\end{equation}
Because $r{=}64 \ll C$, the residual formulation confines the adaptation to a low-rank subspace, so that $\mathcal{F}$ acts as a correction rather than a complete feature rewrite. The multi-scale depthwise convolutions supply spatially adaptive context at three receptive-field sizes without the quadratic cost of a second attention layer. This design distinguishes ASRM from standard gated normalization, which lacks a spatial adaptation path, and from adapter-style modules, which lack the gated normalization front-end that governs the identity-to-regulation transition.

CSIF uses $n{=}2$ CSIB blocks at $P_5$, balancing modeling capacity against the over-smoothing that additional blocks would introduce at $20{\times}20$ resolution. Through global attention, the module allows direct interaction between RGB-derived and depth-derived features across all spatial positions, enabling the network to learn cross-modal correspondences useful for defect detection. The refined $P_5$ features then pass through the terminal SRM for a final round of bottleneck-based distribution normalization before entering the detection head.

\section{Experiments}
\label{sec:experiments}
\subsection{Datasets}
\label{sec:dataset}
This study employs the TL-RGBD (Transmission Line RGB-D Dataset), a large-scale, high-resolution RGB-D dataset specifically designed for detecting defects in power transmission line components. The dataset comprises 10,000 paired RGB and depth images of insulators, tie wires, and poles, captured under authentic inspection scenarios by China Southern Power Grid using UAVs equipped with synchronized RGB cameras and depth sensors. The depth information, acquired directly through onboard depth cameras during flight missions, provides complementary geometric and spatial cues essential for precise defect localization.

Each image pair (RGB and depth) was annotated by domain experts with bounding boxes and category labels, ensuring high-quality and consistent labeling. The dataset encompasses 73,448 annotated instances across nine component states and defect types, as detailed in Table~\ref{tab:label_category}.

\begin{table}[h]
	\centering
	\caption{Correspondence between labels and category names.}
	\label{tab:label_category}
	\setlength{\tabcolsep}{6pt}
	\begin{tabular*}{\columnwidth}
		{@{\hspace{6pt}}c@{\extracolsep{\fill}}l@{\extracolsep{\fill}}l@{\hspace{6pt}}}
		\toprule
		No. & Label & Category name \\
		\midrule
		1 & zcjyz & Normal insulator \\
		2 & jyzwh & Polluted insulator \\
		3 & dgss  & Damaged pole \\
		4 & zxqs  & Missing tie wire \\
		5 & zxst  & Loose tie wire \\
		6 & jyzsl & Insulator flashover \\
		7 & nw    & Bird nest \\
		8 & jyzps & Broken insulator \\
		9 & zyzsl & Shattered post insulator \\
		\bottomrule
	\end{tabular*}
\end{table}

The TL-RGBD dataset exhibits several distinctive characteristics that reflect the intrinsic challenges of UAV-based transmission line inspection. Statistical analysis reveals a highly imbalanced distribution of object sizes, with each annotated instance occupying an average area of 422.12 pixels. The dataset composition is dominated by small-scale targets: 94.51\% of instances are classified as small objects (69,413 instances with area $< 32^{2}$ pixels), 5.48\% as medium-sized objects (4,022 instances with $32^{2} \leq$ area $< 96^{2}$ pixels), and merely 0.02\% as large objects (13 instances with area $\geq 96^{2}$ pixels). This extreme skew toward small-scale targets poses significant challenges for detection algorithms and faithfully represents real-world inspection scenarios where defects often manifest as subtle, localized anomalies.

For experimental evaluation, the TL-RGBD dataset was randomly divided into training, validation, and test subsets following an 8:1:1 split ratio. 80\% of the image pairs were used for model training, 10\% for validation and hyperparameter selection, and the remaining 10\% were reserved for final performance evaluation. The split was conducted at the image level to ensure that paired RGB and depth images always reside within the same subset. This partitioning strategy provides sufficient training data while enabling unbiased validation and testing under consistent data distributions.

The depth maps, captured synchronously with RGB images using onboard depth sensors, encode critical geometric cues that complement RGB appearance features. Depth data is particularly valuable for distinguishing between visually similar defects, identifying component protrusions or deformations, quantifying geometric anomalies, and improving localization accuracy under challenging illumination conditions where RGB information alone may be insufficient. This multimodal approach enables more robust defect characterization by leveraging both photometric and geometric properties of transmission line components.

\begin{figure}[h]
	\centering
	\begin{subfigure}{0.48\columnwidth}
		\includegraphics[width=\linewidth]{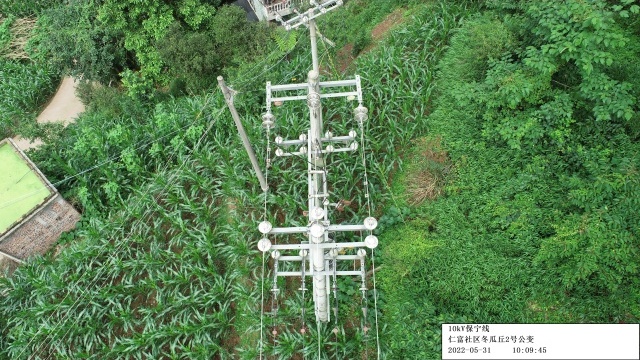}
		\caption{}
	\end{subfigure}
	\hfill
	\begin{subfigure}{0.48\columnwidth}
		\includegraphics[width=\linewidth]{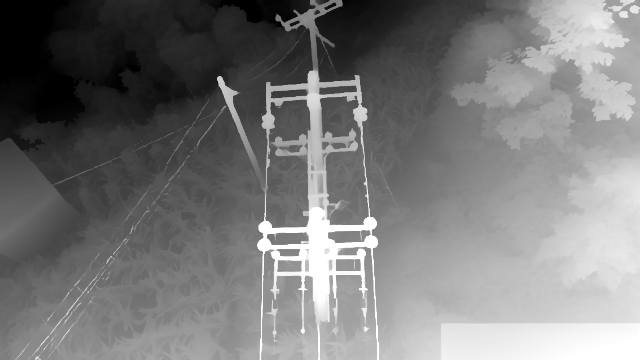}
		\caption{}
	\end{subfigure}
	
	\vspace{0.5em}
	
	\begin{subfigure}{0.48\columnwidth}
		\includegraphics[width=\linewidth]{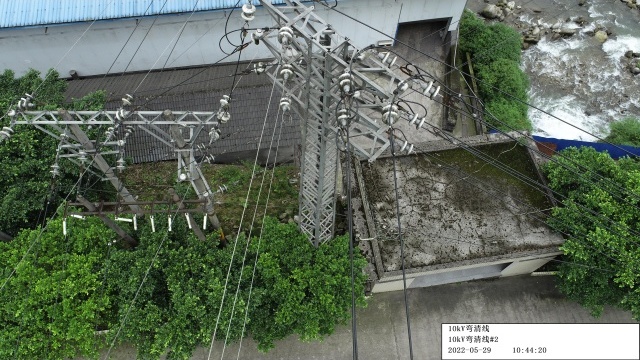}
		\caption{}
	\end{subfigure}
	\hfill
	\begin{subfigure}{0.48\columnwidth}
		\includegraphics[width=\linewidth]{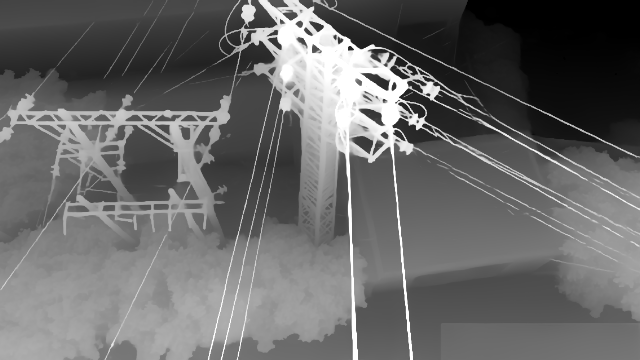}
		\caption{}
	\end{subfigure}
	\caption{Sample image pairs from the TL-RGBD dataset. (a) and (c) are RGB images captured by UAV-mounted cameras showing transmission line components under real inspection conditions. (b) and (d) are the corresponding depth maps acquired synchronously, providing geometric information for defect localization.}
	\label{fig:dataset_samples}
\end{figure}

\begin{figure}[h]
	\centering
	\includegraphics[width=0.85\columnwidth]{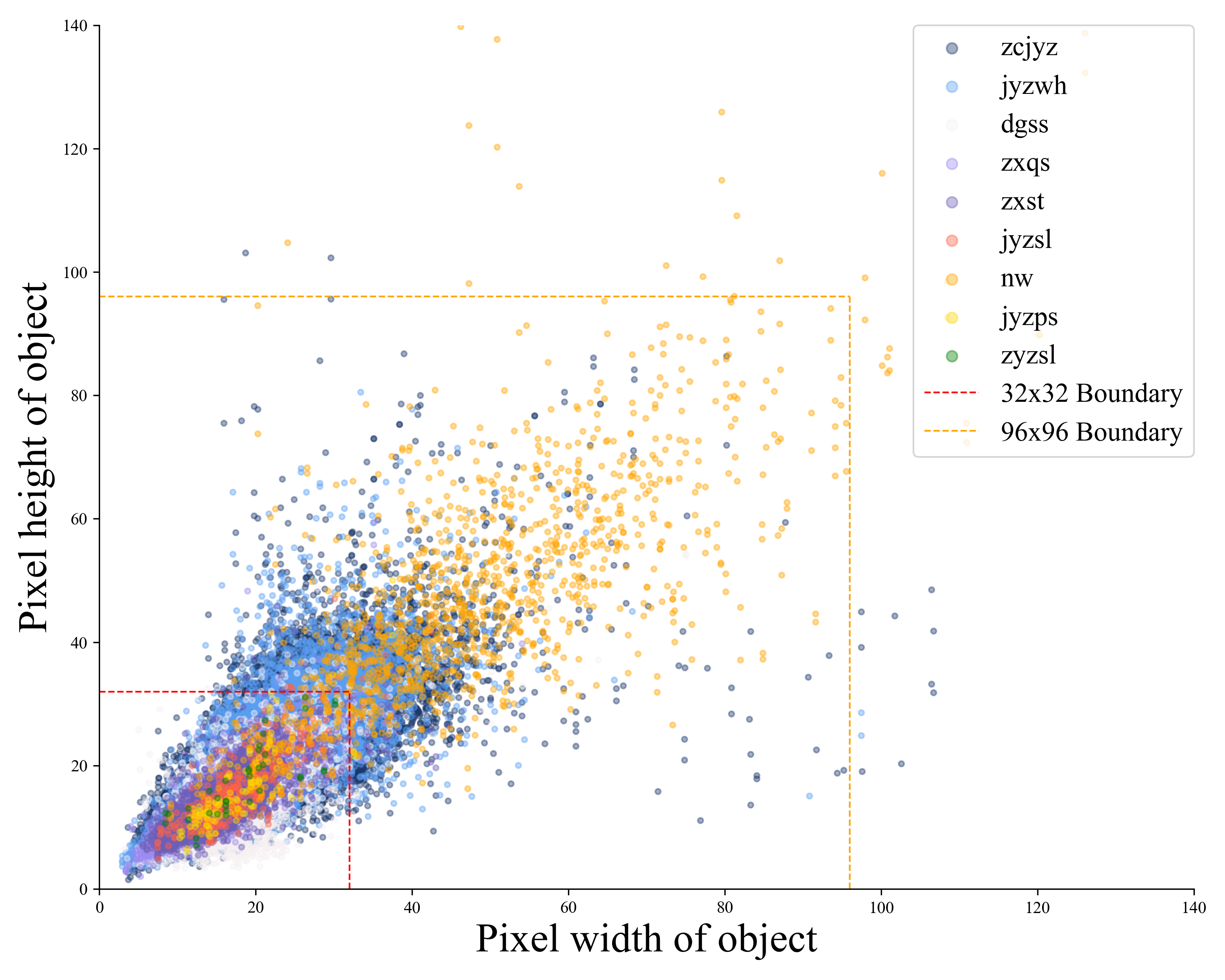}
	\caption{Distribution of object sizes in the TL-RGBD dataset. The scatter plot shows the pixel width and height of all annotated instances after preprocessing, with different colors representing distinct defect categories. The concentration of points in the lower-left region indicates the predominance of small-scale objects. Dashed lines at 32$\times$32 and 96$\times$96 pixels mark the boundaries between small, medium, and large object categories.}
	\label{fig:Distribution}
\end{figure}

In summary, the TL-RGBD dataset with native RGB-D modalities provides a robust, high-fidelity benchmark that advances both domain-specific defect detection research and general small-object detection methodologies in aerial remote sensing applications. The combination of authentic inspection data with synchronized depth sensing offers enhanced domain relevance and addresses challenging detection scenarios specific to critical infrastructure monitoring.

\subsection{Model Training and Evaluation Metrics}

The proposed cross-modal alignment and fusion network was implemented using the PyTorch framework (version 2.3.1). All experiments were conducted on a workstation equipped with an NVIDIA GeForce RTX 3090 GPU (24 GB memory) running Ubuntu 22.04 LTS with CUDA 12.1 and Python 3.10.14. During training, a batch size of 16 was adopted for all experiments, balancing GPU memory constraints and optimization stability on the RTX 3090 platform.

Model performance was assessed on the held-out test set following the standard COCO evaluation protocol. We report two primary metrics: mAP$_{50}$, which measures mean average precision at an intersection-over-union (IoU) threshold of 0.50, and mAP$_{50:95}$, which averages precision across IoU thresholds ranging from 0.50 to 0.95 in increments of 0.05.

Given the prevalence of small-scale defects in transmission-line inspection scenarios captured by UAV-mounted RGB-D sensors, we additionally report scale-sensitive metrics defined by the COCO protocol: $\text{AP}_s$ for small objects (area $< 32^2$ pixels) and $\text{AP}_m$ for medium objects (area between $32^2$ and $96^2$ pixels). The $\text{AP}_s$ metric is particularly informative for assessing the capability of a model to detect fine-grained, low-pixel-count anomalies---a critical requirement for reliable transmission-line monitoring. Computational efficiency and model complexity were quantified using floating-point operations (FLOPs) and the total number of parameters (Params, in millions). Detection performance was evaluated through precision ($P$), recall ($R$), average precision (AP), and mean average precision (mAP), defined as follows:
\[
P = \frac{TP}{TP + FP}, \qquad R = \frac{TP}{TP + FN},
\]
where $TP$, $FP$, and $FN$ denote true positives, false positives, and false negatives, respectively. For each defect category, the average precision is computed as the area under the precision-recall curve:
\[
\text{AP} = \int_{0}^{1} P(R) \, dR,
\]
and mean average precision across $k$ categories is given by:
\[
\text{mAP} = \frac{1}{k} \sum_{i=1}^{k} \text{AP}_i.
\]

Collectively, these computational metrics (FLOPs, Params) and detection metrics (mAP, scale-sensitive AP) provide a balanced evaluation of accuracy, completeness, and operational efficiency for RGB-D sensor-based transmission-line defect detection.

\subsection{Model Scaling Strategy}
\label{sec:scaling}

CMAFNet adopts a unified scaling paradigm that enables systematic adaptation across diverse deployment scenarios without modifying the underlying network topology. Following the compound scaling methodology established in modern detection frameworks, the architecture defines five capacity configurations---nano (n), small (s), medium (m), large (l), and extra-large (x)---through coordinated adjustment of three scaling coefficients. The \textit{depth\_multiple} coefficient governs the repetition count of stackable blocks within each stage, the \textit{width\_multiple} coefficient controls the channel dimensionality at each layer, and the \textit{max\_channels} parameter imposes an upper bound on deep-layer channel expansion to prevent computational overhead from scaling superlinearly in dual-branch fusion architectures. This parameterization produces a family of models that share identical architectural topology and module composition, differing only in representational capacity and computational cost.

Table~\ref{tab:scaling_config} summarizes the scaling coefficients for each configuration. The transitions between adjacent scales embody distinct capacity expansion strategies. The nano-to-small transition increases channel width while preserving depth, thereby expanding feature dimensionality for richer local representations. The small-to-medium transition further widens channels to baseline capacity while introducing deep-layer channel capping to constrain computation at the fusion stage. The medium-to-large transition restores full block repetition counts to enhance the depth of hierarchical abstraction. The large-to-extra-large transition extends mid-to-high-level channel width beyond the baseline under continued deep-layer capping, maximizing the capacity for cross-modal fusion.

\begin{table}[h]
	\centering
	\caption{Scaling configurations of CMAFNet variants.}
	\label{tab:scaling_config}
	\footnotesize
	\begin{tabular*}{\columnwidth}{
			@{\hspace{2pt}}l
			@{\extracolsep{\fill}}ccc
			@{\hspace{2pt}}
		}
		\toprule
		Scale & Depth & Width & Max Channels \\
		\midrule
		Nano (n)        & 0.50 & 0.25 & 1024 \\
		Small (s)       & 0.50 & 0.50 & 1024 \\
		Medium (m)      & 0.50 & 1.00 & 512  \\
		Large (l)       & 1.00 & 1.00 & 512  \\
		Extra-large (x) & 1.00 & 1.50 & 512  \\
		\bottomrule
	\end{tabular*}
\end{table}

The rationale for incorporating the \textit{max\_channels} constraint merits attention in the context of dual-branch RGB-D architectures. Unlike single-stream detectors where channel expansion propagates uniformly through the network, the proposed fusion strategy concatenates features from parallel branches at $P_4$ and $P_5$ scales, effectively doubling the channel count at fusion points before subsequent compression. Without explicit bounds, applying large width multipliers would cause deep-layer channels to exceed practical memory limits, particularly at the $P_5$ fusion stage where the Contextual Semantic Integration Framework performs global attention operations with quadratic memory complexity in the channel dimension. Setting \textit{max\_channels} to 512 for medium through extra-large configurations ensures that channels at the fusion stage remain tractable while permitting aggressive width expansion at shallower scales where cross-modal concatenation has not yet occurred.

The selection of an appropriate model scale should be guided by a hierarchical decision framework that prioritizes task-side constraints over deployment-side constraints when both are applicable. Transmission-line defect detection presents a canonical example of task-driven scaling requirements: with 94.51\% of annotated instances in the TL-RGBD dataset classified as small objects under COCO criteria, the detection task demands high-capacity models capable of preserving fine-grained features through deep pyramid pathways while accommodating the expanded representational requirements of cross-modal fusion. Small targets impose stringent demands on feature resolution and semantic granularity---characteristics that lightweight models struggle to maintain as features propagate through successive downsampling and fusion operations. When the application domain exhibits predominantly small-object distributions or requires high localization precision for subtle defects, large or extra-large configurations are recommended regardless of available computational resources.

When task-side constraints do not unambiguously dictate capacity requirements---for instance, in scenarios involving mixed object scales or moderate precision tolerances---the deployment platform becomes the primary determinant of model selection. Edge devices such as UAV-mounted inference units or embedded processors impose severe constraints on power consumption, memory bandwidth, and computational throughput, necessitating the nano configuration to achieve stable real-time inference despite moderate accuracy reduction. Mobile computing platforms including portable inspection terminals and field workstations offer intermediate computational capacity, rendering the small configuration appropriate for balancing responsiveness with detection fidelity. Desktop GPUs and server-class accelerators provide sufficient resources to support medium through extra-large configurations, with the specific choice determined by latency requirements and operational modes. For batch processing of archived inspection imagery where throughput rather than latency governs efficiency, the extra-large configuration maximizes detection accuracy without deployment constraints.

This task-platform scaling framework yields consistent model selection across application scenarios: small-object-dominated tasks direct selection toward high-capacity configurations, while resource-constrained deployments direct selection toward lightweight configurations. The framework resolves potential conflicts by assigning precedence to task requirements---if small-object detection mandates high capacity but edge deployment requires low complexity, practitioners must either accept reduced accuracy on edge platforms or migrate inference to more capable hardware. The experimental results presented in Table~\ref{tab:full_model_comparison} validate this framework empirically: the monotonic relationship between model capacity and small-object performance across all five scales confirms that the scaling strategy successfully translates architectural capacity into detection capability for the target application domain.

\subsection{Comparisons with Previous Methods}

We benchmark CMAFNet against state-of-the-art object detection methods spanning three architectural paradigms: CNN-based single-stage detectors from the YOLO family (v5, v8, v10, v11), transformer-based end-to-end detectors built on the DETR framework (RT-DETR, Conditional-DETR, DETR, DINO), and a specialized small-object detector (FFCA-YOLO). Additionally, we compare against TinyDef-DETR, a domain-specific transformer tailored for transmission-line defect detection. For RGB-only baselines, we adopt results from our prior work~\citep{shenTinyDefDETRTransformerBasedFramework2025}, ensuring identical dataset splits and training configurations for fair comparison.

For all compared detectors, we followed the official training recipes, that is, the default hyperparameter settings recommended in the corresponding open-source implementations or described in the original papers. To avoid unintentionally biasing the comparison through dataset-specific hyperparameter tuning, we did not modify algorithm-specific optimization strategies, including the optimizer type, learning-rate schedule, or assignment and matching rules. Only dataset-dependent adaptations were applied when strictly necessary, such as adjusting the number of categories and data loading paths.

Unless otherwise specified, all methods were trained and evaluated using the same input resolution and the same TL-RGBD train/validation/test split, ensuring a fair and controlled comparison across different architectural paradigms.

Table~\ref{tab:full_model_comparison} presents the quantitative results on the TL-RGBD dataset. CMAFNet-x achieves the highest mAP$_{50}$ of 0.322, surpassing the best-performing baseline DINO (0.224) by 9.8 percentage points. This superiority persists across all model scales: CMAFNet-n attains an mAP$_{50}$ of 0.248 with only 4.9M parameters and 12.4 GFLOPs, outperforming all YOLO variants and matching larger transformer models. Such scaling behavior indicates that the performance gains originate primarily from cross-modal fusion rather than from increased model capacity alone. For small-object detection---the core challenge in transmission-line imagery---CMAFNet-x achieves an AP$_s$ of 0.125, improving upon DINO (0.085) by 4.0 percentage points and substantially exceeding YOLO variants (0.06--0.08). The mAP$_{50:95}$ metric further confirms this advantage: CMAFNet-x reaches 0.147 versus 0.099 for DINO and 0.072--0.083 for leading YOLO configurations, demonstrating that cross-modal fusion enhances both detection recall and boundary localization. Although large objects constitute merely 0.02\% of the TL-RGBD dataset (13 instances with area $\geq 96^{2}$ pixels), we report AP$_l$ for completeness; notably, even the smallest variant CMAFNet-n achieves 0.369 on this subset, providing a conservative reference point that larger configurations are expected to surpass. Category-specific metrics for \textit{jyzwh} and \textit{dgss}---the two most prevalent defect types in the TL-RGBD dataset---reveal pronounced improvements: CMAFNet-x achieves a recall of 0.584 and mAP$_{50}$ of 0.517 for \textit{jyzwh} defects (DINO: 0.437 and 0.420), and a recall of 0.463 with mAP$_{50}$ of 0.465 for \textit{dgss} defects (DINO: 0.402 and 0.426). These results validate that the depth modality provides discriminative geometric cues complementing RGB appearance features, particularly for defects with subtle visual signatures but distinctive three-dimensional profiles.

\begin{table*}[h]
	\centering
	\caption{Comprehensive performance comparison of different object detection models on the TL-RGBD dataset. The best results are highlighted in bold.}
	\label{tab:full_model_comparison}
	\footnotesize
	\setlength{\tabcolsep}{2.4pt}
	\begin{tabular*}{\textwidth}{
			@{\hspace{3pt}}l
			@{\extracolsep{\fill}}
			c c c c c c c c c c c c c c
			@{\hspace{3pt}}
		}
		\toprule
		Model & Epochs & Params(M) & GFLOPs & P & R & mAP$_{50}$ & mAP$_{50:95}$ & AP$_s$ & AP$_m$ & FPS & Recall$_{jyzwh}$ & Recall$_{dgss}$ & mAP$_{50}^{jyzwh}$ & mAP$_{50}^{dgss}$ \\
		\midrule
		YOLO v5n & 300 & \textbf{2.2} & \textbf{5.8} & \textbf{0.707} & 0.131 & 0.131 & 0.063 & 0.061 & 0.097 & 243.9 & 0.352 & 0.292 & 0.315 & 0.258 \\
		YOLO v5s & 300 & 9.1 & 23.8 & 0.346 & 0.148 & 0.140 & 0.065 & 0.065 & 0.098 & 181.8 & 0.377 & 0.323 & 0.314 & 0.270 \\
		YOLO v5m & 300 & 25.1 & 64.0 & 0.548 & 0.164 & 0.162 & 0.072 & 0.079 & 0.081 & 74.6 & 0.372 & 0.321 & 0.367 & 0.277 \\
		YOLO v8n & 300 & 2.7 & 6.8 & 0.580 & 0.134 & 0.133 & 0.061 & 0.066 & 0.078 & 263.2 & 0.364 & 0.303 & 0.321 & 0.264 \\
		YOLO v8s & 300 & 11.1 & 28.5 & 0.375 & 0.155 & 0.147 & 0.066 & 0.067 & 0.095 & 147.1 & 0.388 & 0.321 & 0.331 & 0.269 \\
		YOLO v8m & 300 & 23.2 & 67.5 & 0.403 & 0.169 & 0.163 & 0.072 & 0.075 & 0.087 & 66.2 & 0.378 & 0.328 & 0.360 & 0.296 \\
		YOLO v10n & 300 & 2.7 & 8.2 & \textbf{0.707} & 0.125 & 0.133 & 0.063 & 0.062 & 0.094 & \textbf{270.3} & 0.352 & 0.237 & 0.321 & 0.251 \\
		YOLO v10s & 300 & 8.0 & 24.5 & 0.536 & 0.144 & 0.157 & 0.073 & 0.078 & 0.095 & 140.8 & 0.343 & 0.288 & 0.341 & 0.308 \\
		YOLO v10m & 300 & 16.6 & 63.5 & 0.386 & 0.170 & 0.162 & 0.075 & 0.075 & 0.097 & 67.1 & 0.390 & 0.361 & 0.356 & 0.294 \\
		YOLO 11n & 300 & 2.6 & 6.3 & 0.638 & 0.135 & 0.136 & 0.064 & 0.060 & 0.103 & 263.2 & 0.353 & 0.274 & 0.324 & 0.245 \\
		YOLO 11s & 300 & 9.4 & 21.3 & 0.516 & 0.166 & 0.155 & 0.072 & 0.074 & 0.100 & 125.0 & 0.381 & 0.314 & 0.344 & 0.291 \\
		YOLO 11s-P2 & 300 & 9.6 & 28.7 & 0.608 & 0.183 & 0.181 & 0.083 & 0.087 & 0.100 & 99.0 & 0.384 & 0.326 & 0.366 & 0.323 \\
		YOLO 11m & 300 & 20.0 & 67.7 & 0.399 & 0.181 & 0.173 & 0.078 & 0.082 & 0.094 & 65.8 & 0.382 & 0.358 & 0.373 & 0.307 \\
		RT-DETR-l & 100 & 32.8 & 108.0 & 0.480 & 0.189 & 0.199 & 0.089 & 0.078 & 0.044 & 38.5 & 0.360 & 0.337 & 0.341 & 0.372 \\
		RT-DETR-x & 100 & 65.5 & 222.5 & 0.485 & 0.199 & 0.216 & 0.099 & 0.083 & 0.058 & 14.9 & 0.366 & 0.371 & 0.363 & 0.403 \\
		RT-DETR-R18 & 100 & 19.8 & 57.0 & 0.369 & 0.177 & 0.163 & 0.074 & 0.071 & 0.039 & 88.0 & 0.347 & 0.381 & 0.316 & 0.354 \\
		TinyDef-DETR & 100 & 20.5 & 65.3 & 0.534 & 0.263 & 0.275 & 0.119 & 0.106 & \textbf{0.110} & 62.8 & 0.397 & 0.356 & 0.391 & \textbf{0.419} \\
		RT-DETR-R34 & 100 & 31.2 & 88.8 & 0.354 & 0.127 & 0.117 & 0.053 & 0.056 & 0.062 & 61.1 & 0.293 & 0.338 & 0.260 & 0.284 \\
		RT-DETR-R50 & 100 & 42.9 & 134.8 & 0.363 & 0.120 & 0.115 & 0.051 & 0.054 & 0.027 & 36.5 & 0.278 & 0.292 & 0.254 & 0.275 \\
		RT-DETR-R101 & 100 & 74.7 & 247.1 & 0.467 & 0.185 & 0.193 & 0.088 & 0.073 & 0.050 & 25.6 & 0.345 & 0.367 & 0.316 & 0.378 \\
		Conditional-DETR-R18 & 50 & --- & --- & 0.309 & 0.068 & 0.038 & 0.011 & 0.009 & 0.016 & --- & 0.150 & 0.111 & 0.073 & 0.082 \\
		Conditional-DETR-R34 & 50 & --- & --- & 0.447 & 0.082 & 0.034 & 0.009 & 0.006 & 0.011 & --- & 0.184 & 0.123 & 0.063 & 0.059 \\
		Conditional-DETR-R50 & 50 & 44.0 & 89.5 & 0.568 & 0.159 & 0.172 & 0.066 & 0.064 & 0.057 & 41.1 & 0.406 & 0.294 & 0.389 & 0.287 \\
		DETR-R50 & 500 & --- & --- & 0.301 & 0.138 & 0.124 & 0.043 & 0.041 & 0.037 & --- & 0.289 & 0.258 & 0.280 & 0.203 \\
		DINO & 12 & --- & --- & 0.496 & 0.257 & 0.224 & 0.099 & 0.085 & 0.075 & --- & 0.437 & 0.402 & 0.420 & 0.426 \\
		FFCA-YOLO & 300 & 7.1 & 51.4 & 0.680 & 0.130 & 0.142 & 0.060 & 0.059 & 0.051 & 50.8 & 0.286 & 0.291 & 0.291 & 0.270 \\
		\midrule
		CMAFNet-n & 300 & 4.9 & 12.4 & 0.486 & 0.240 & 0.248 & 0.109 & 0.090 & 0.071 & 227.6 & 0.450 & 0.349 & 0.438 & 0.396 \\
		CMAFNet-s & 300 & 19.7 & 41.8 & 0.502 & 0.281 & 0.270 & 0.121 & 0.099 & 0.068 & 188.6 & 0.496 & 0.420 & 0.456 & 0.420 \\
		CMAFNet-m & 300 & 33.8 & 118.1 & 0.420 & 0.290 & 0.291 & 0.132 & 0.110 & 0.075 & 126.8 & 0.529 & 0.402 & 0.490 & 0.441 \\
		CMAFNet-l & 300 & 42.3 & 149.8 & 0.574 & 0.302 & 0.307 & 0.141 & 0.113 & 0.066 & 96.5 & 0.537 & 0.431 & 0.505 & 0.451 \\
		CMAFNet-x & 300 & 94.3 & 332.2 & 0.418 & \textbf{0.342} & \textbf{0.322} & \textbf{0.147} & \textbf{0.125} & 0.073 & 62.6 & \textbf{0.584} & \textbf{0.463} & \textbf{0.517} & \textbf{0.465} \\
		\bottomrule
	\end{tabular*}
\end{table*}

Analysis of the baseline methods reveals distinct performance patterns across architectural paradigms. Among YOLO detectors, scaling model capacity yields only marginal mAP$_{50}$ improvements of 0.02--0.04 despite substantial increases in parameters and computation. For instance, YOLOv8m (23.2M parameters, 67.5 GFLOPs) achieves an mAP$_{50}$ of 0.163, only 0.030 higher than that of YOLOv8n (2.7M parameters, 6.8 GFLOPs). This saturation suggests that RGB features alone provide limited discriminative power for small, visually ambiguous defects, and that increased network capacity cannot overcome this inherent modality limitation. YOLO 11s-P2, which incorporates an additional high-resolution detection head for small objects, achieves the best YOLO performance (mAP$_{50}$ = 0.181), yet remains substantially below CMAFNet-n despite comparable computational cost. Transformer-based detectors achieve higher accuracy than their YOLO counterparts---DINO reaches 0.224 mAP$_{50}$ and RT-DETR-x reaches 0.216---indicating that global attention mechanisms offer advantages for capturing contextual relationships among spatially distributed defects. However, RT-DETR-x requires 222.5 GFLOPs, approximately 33$\times$ that of YOLOv8n, and performance varies inconsistently across backbones: RT-DETR-R50 (0.115 mAP$_{50}$) underperforms RT-DETR-R18 (0.163 mAP$_{50}$), revealing sensitivity to backbone selection in this domain. FFCA-YOLO, designed specifically for small-object detection, achieves only 0.142 mAP$_{50}$ and 0.059 AP$_s$, underscoring that methods optimized for generic small-object scenarios may not transfer effectively to transmission-line imagery where targets exhibit both small scale and low visual contrast against complex backgrounds.

Comparison with TinyDef-DETR~\citep{shenTinyDefDETRTransformerBasedFramework2025}, representing the prior state of the art for transmission-line defect detection, further demonstrates the effectiveness of cross-modal fusion. Despite the domain-specific architectural refinements in TinyDef-DETR, CMAFNet-x achieves superior performance across all metrics, indicating that geometric cues from depth imagery---when properly aligned---provide discriminative information that single-modal architectural innovations cannot capture. The lightweight CMAFNet-n attains comparable accuracy to TinyDef-DETR with fewer parameters, suggesting that cross-modal fusion offers a more efficient pathway to improved detection than scaling single-modal complexity.

\begin{figure*}[h]
	\centering
	\subfloat[]{%
		\includegraphics[width=0.24\textwidth]{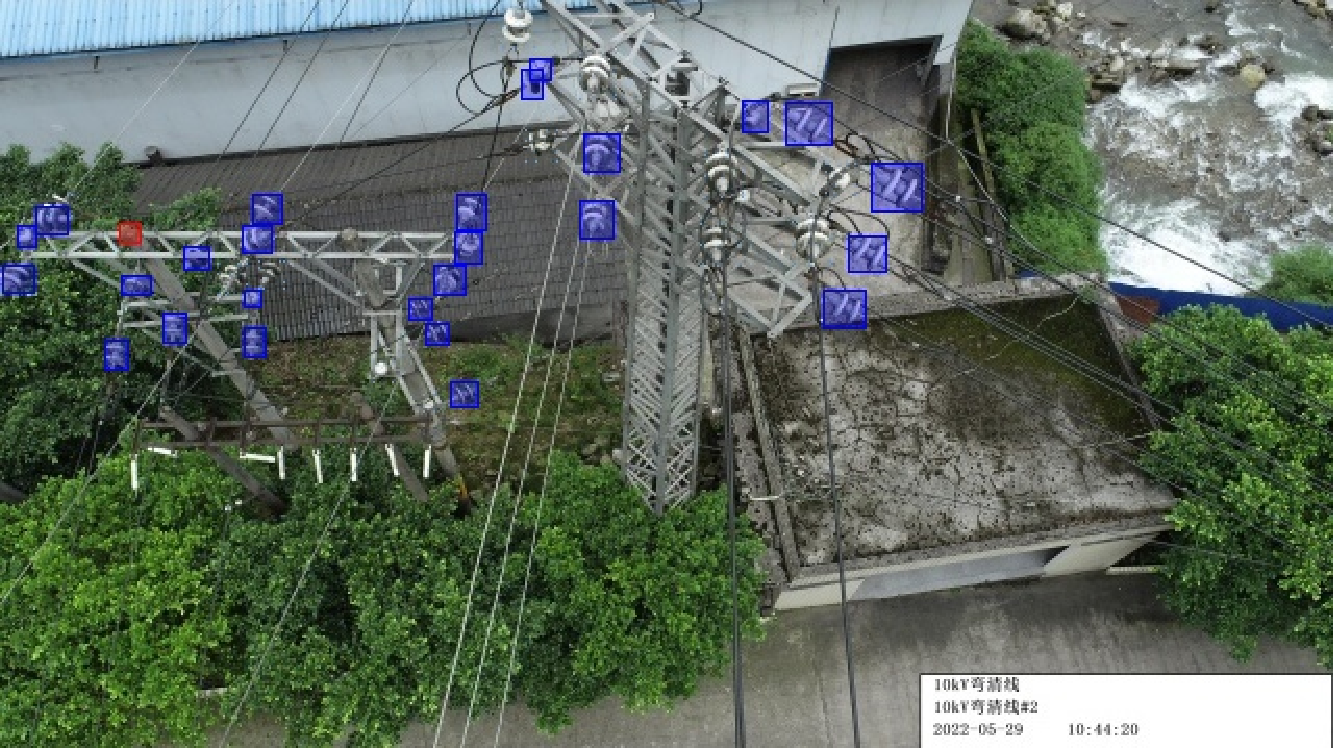}%
	}\hfill
	\subfloat[]{%
		\includegraphics[width=0.24\textwidth]{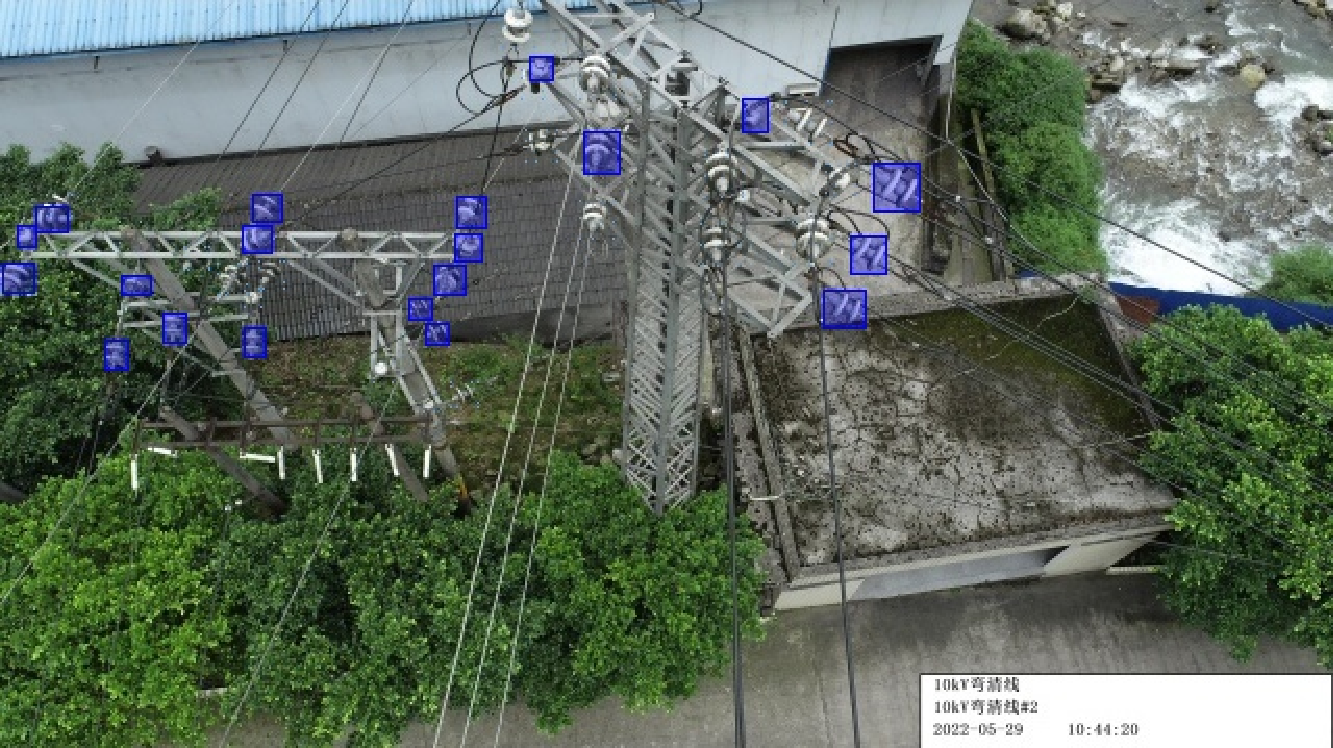}%
	}\hfill
	\subfloat[]{%
		\includegraphics[width=0.24\textwidth]{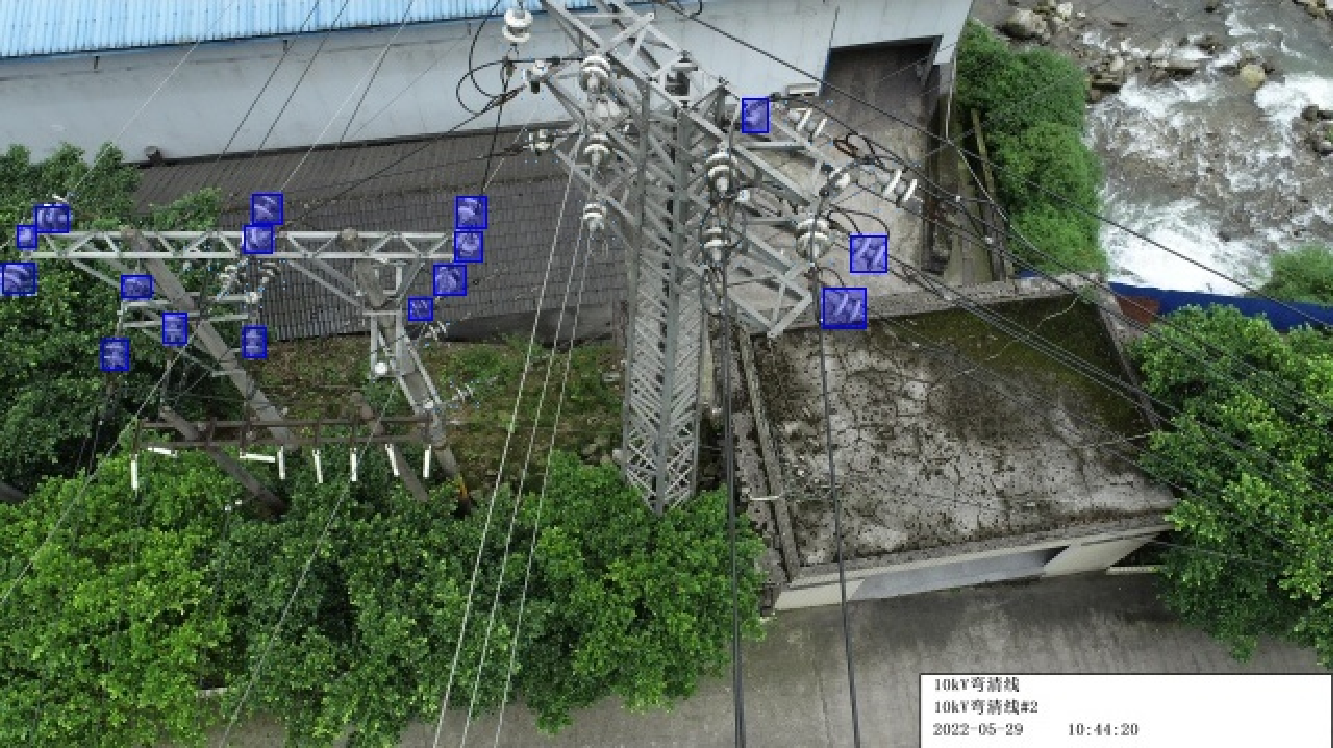}%
	}\hfill
	\subfloat[]{%
		\includegraphics[width=0.24\textwidth]{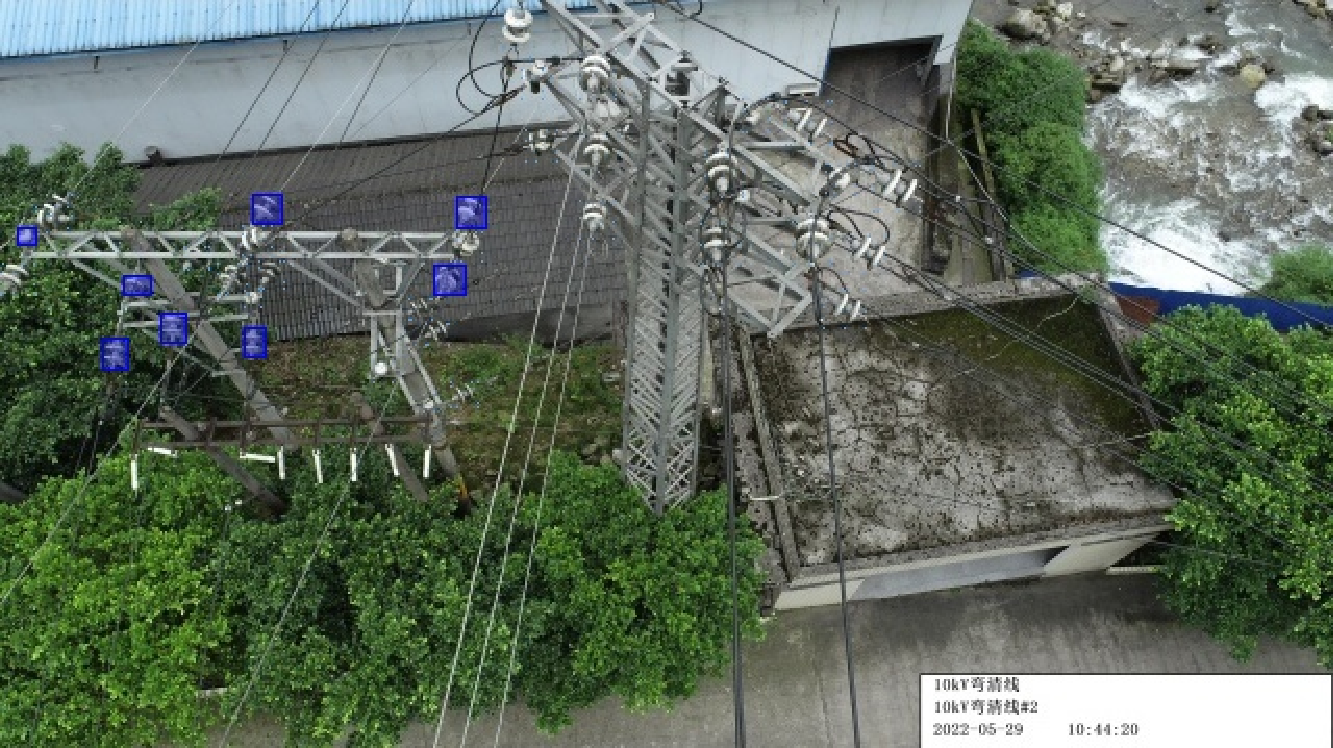}%
	}\\[2pt]
	\subfloat[]{%
		\includegraphics[width=0.24\textwidth]{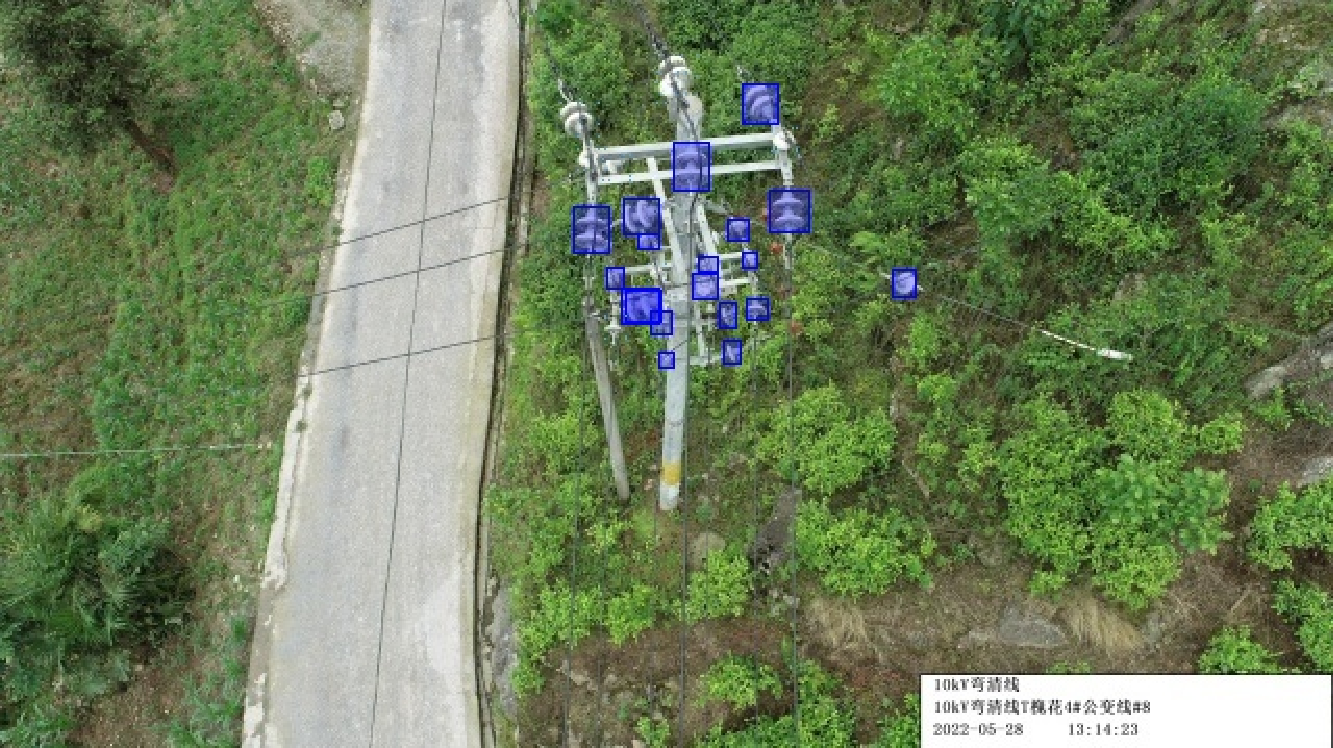}%
	}\hfill
	\subfloat[]{%
		\includegraphics[width=0.24\textwidth]{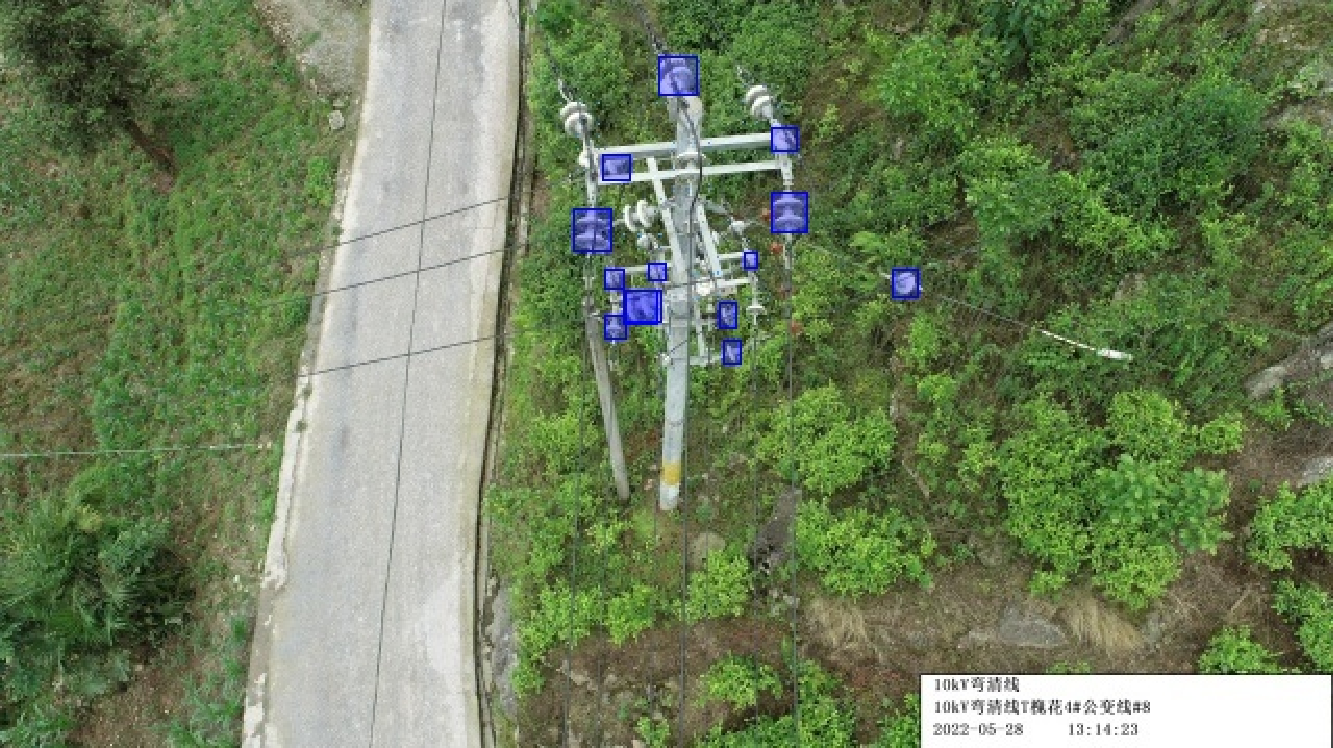}%
	}\hfill
	\subfloat[]{%
		\includegraphics[width=0.24\textwidth]{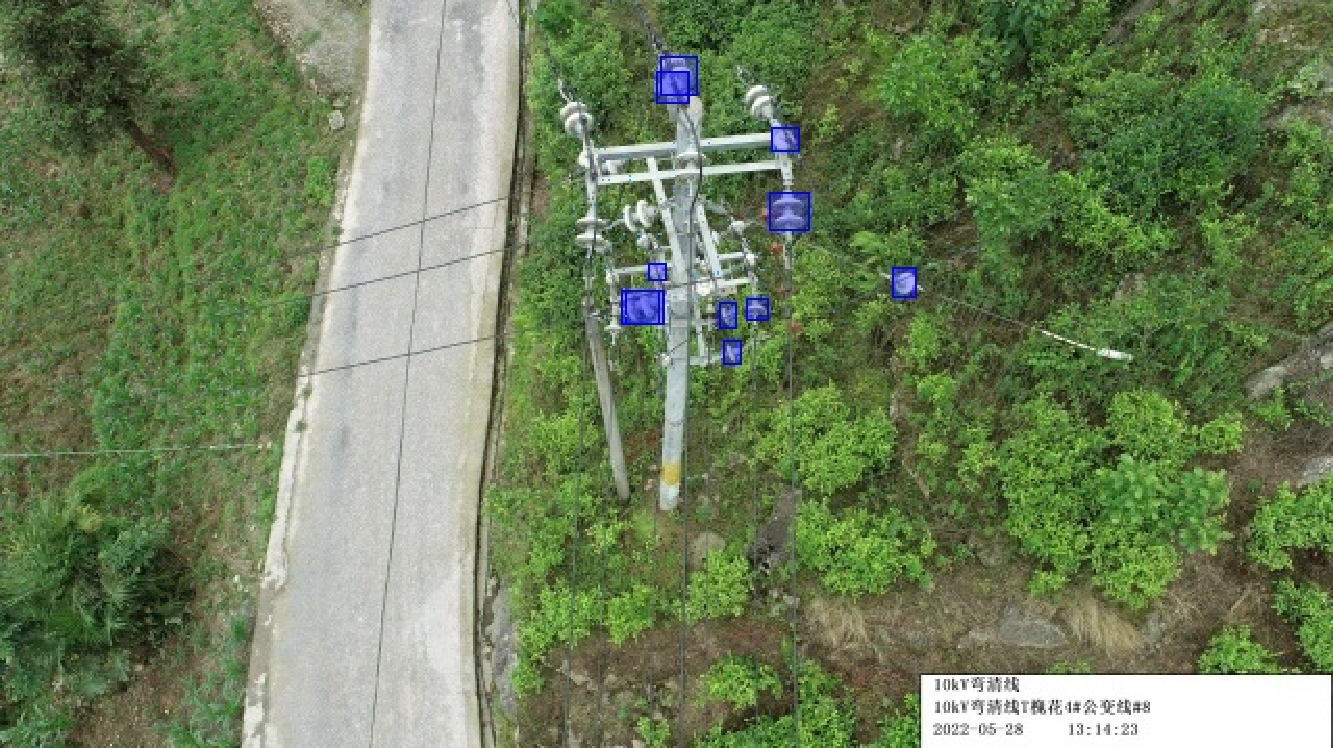}%
	}\hfill
	\subfloat[]{%
		\includegraphics[width=0.24\textwidth]{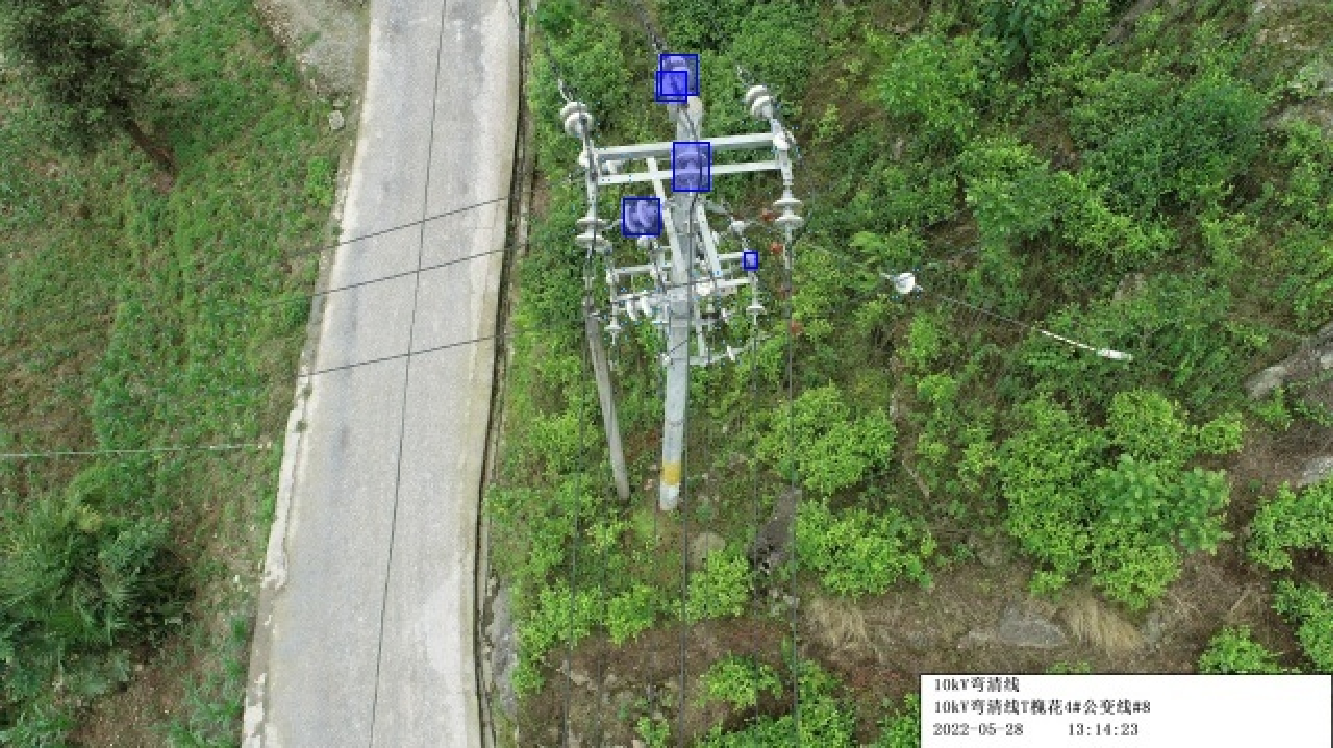}%
	}\\[2pt]
	\subfloat[]{%
		\includegraphics[width=0.24\textwidth]{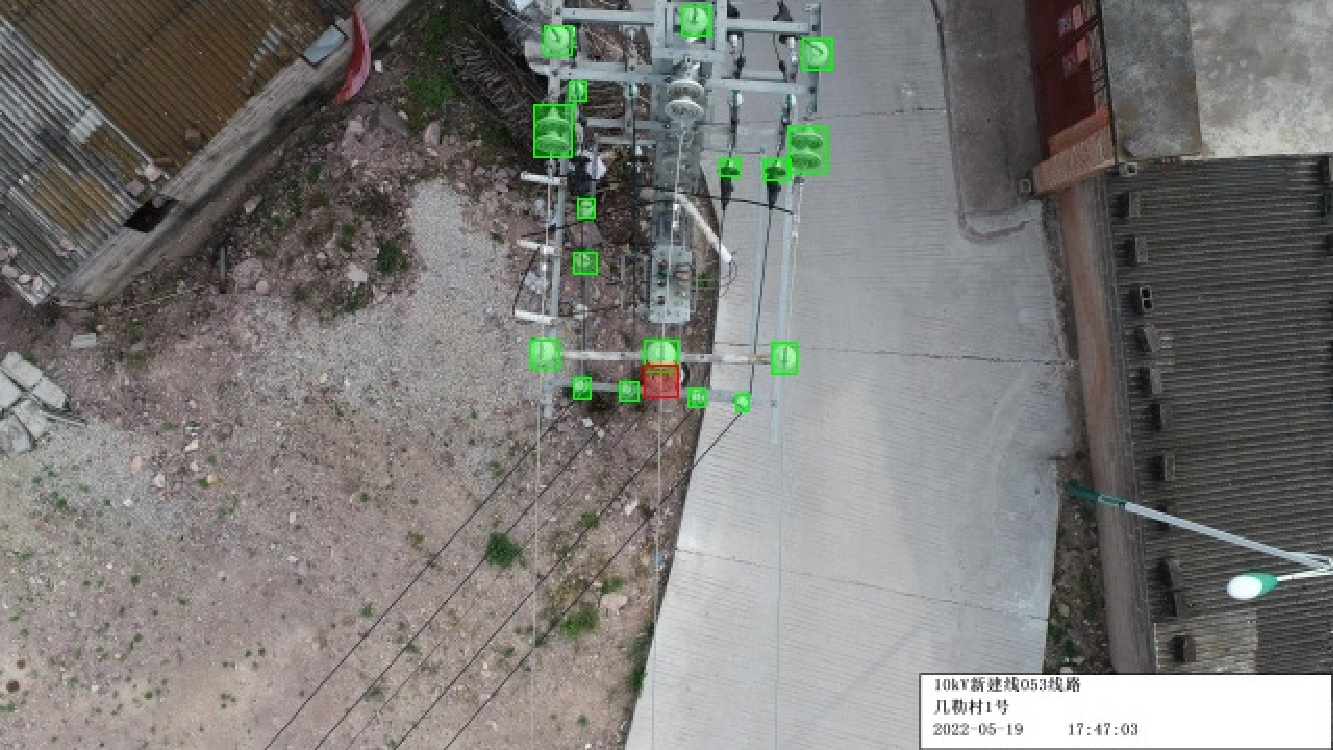}%
	}\hfill
	\subfloat[]{%
		\includegraphics[width=0.24\textwidth]{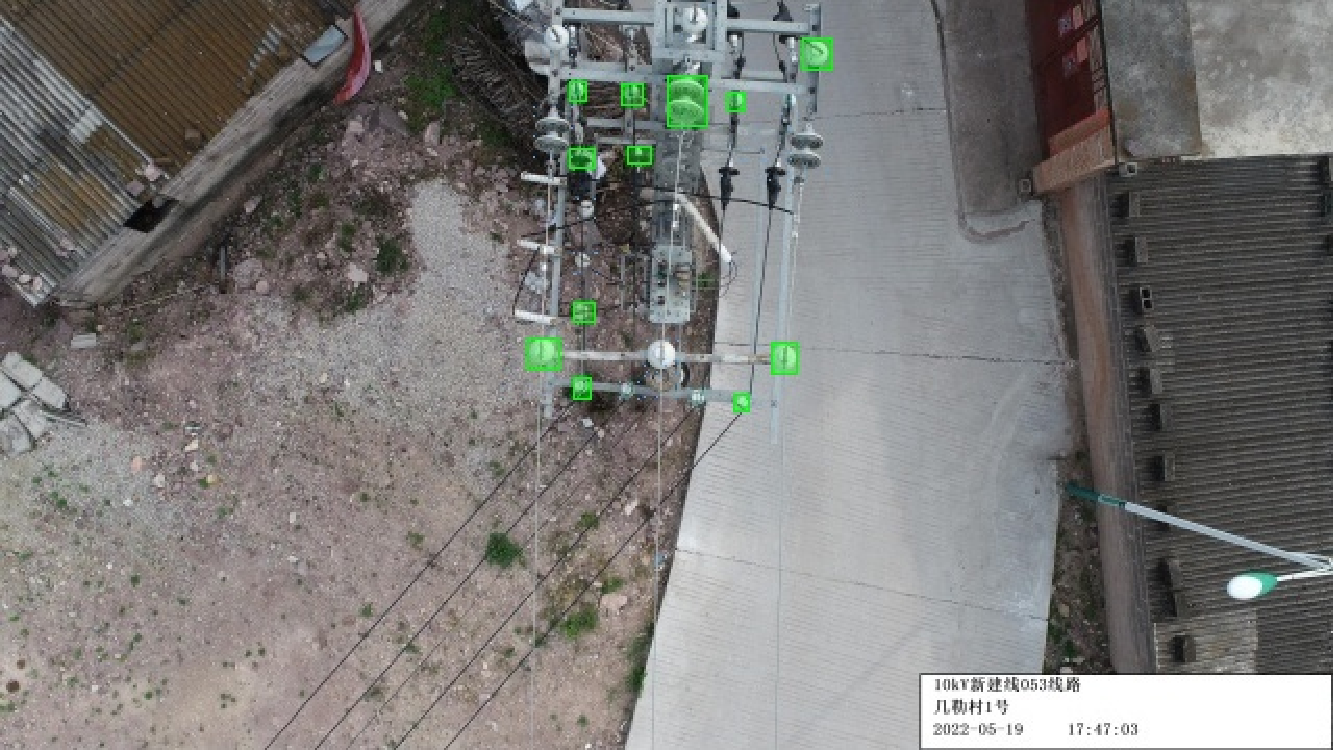}%
	}\hfill
	\subfloat[]{%
		\includegraphics[width=0.24\textwidth]{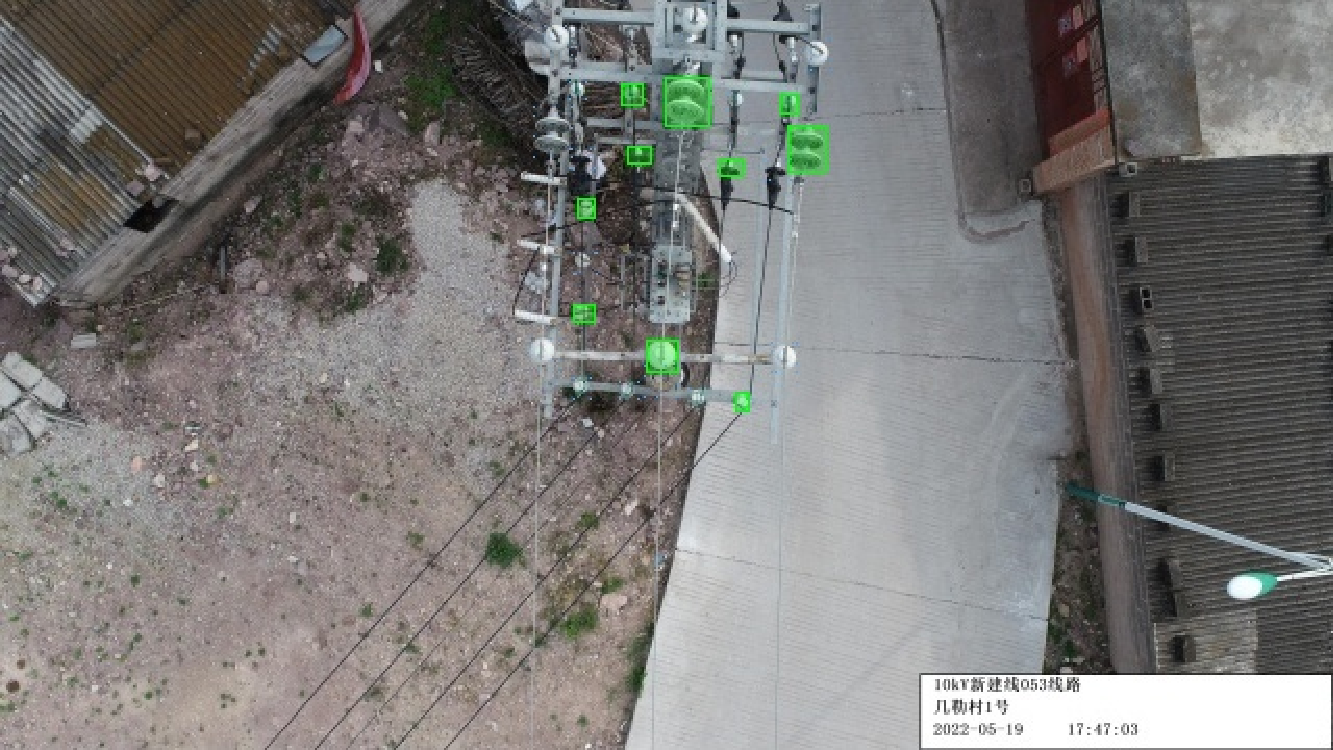}%
	}\hfill
	\subfloat[]{%
		\includegraphics[width=0.24\textwidth]{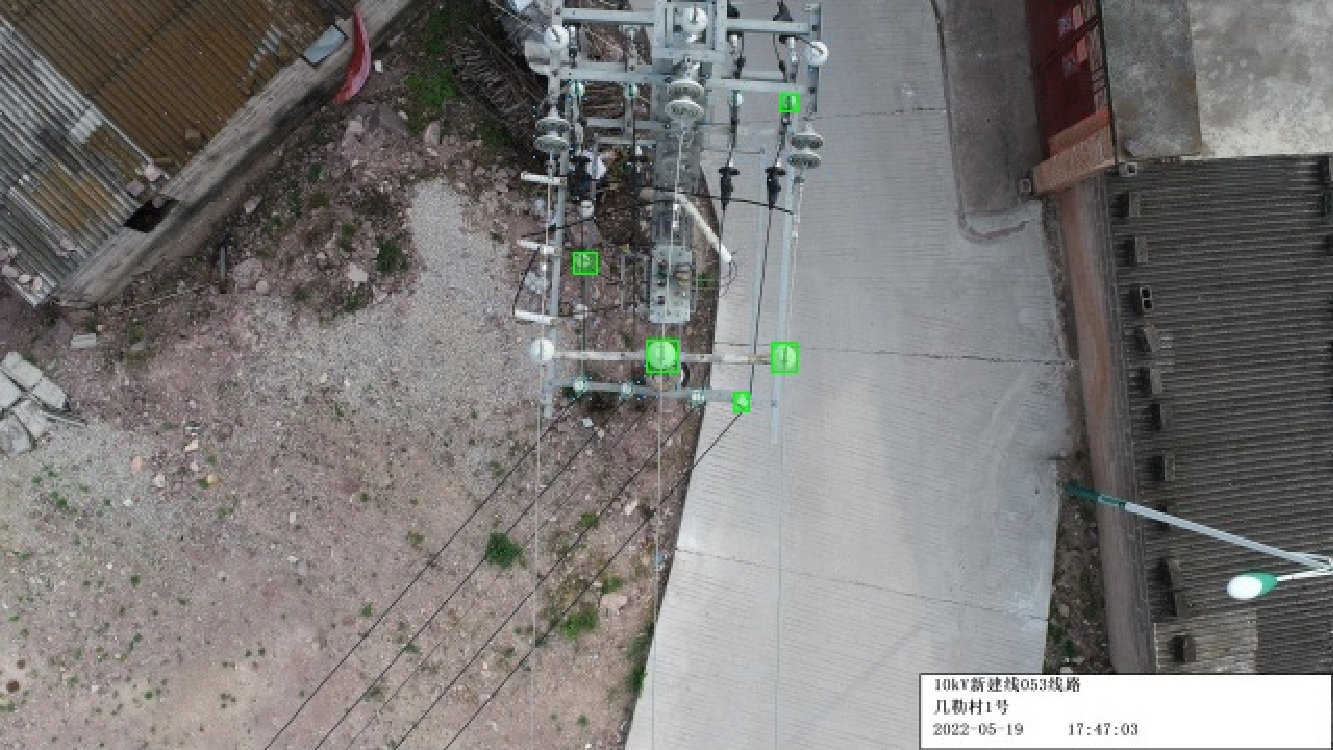}%
	}\\[2pt]
	\subfloat[]{%
		\includegraphics[width=0.24\textwidth]{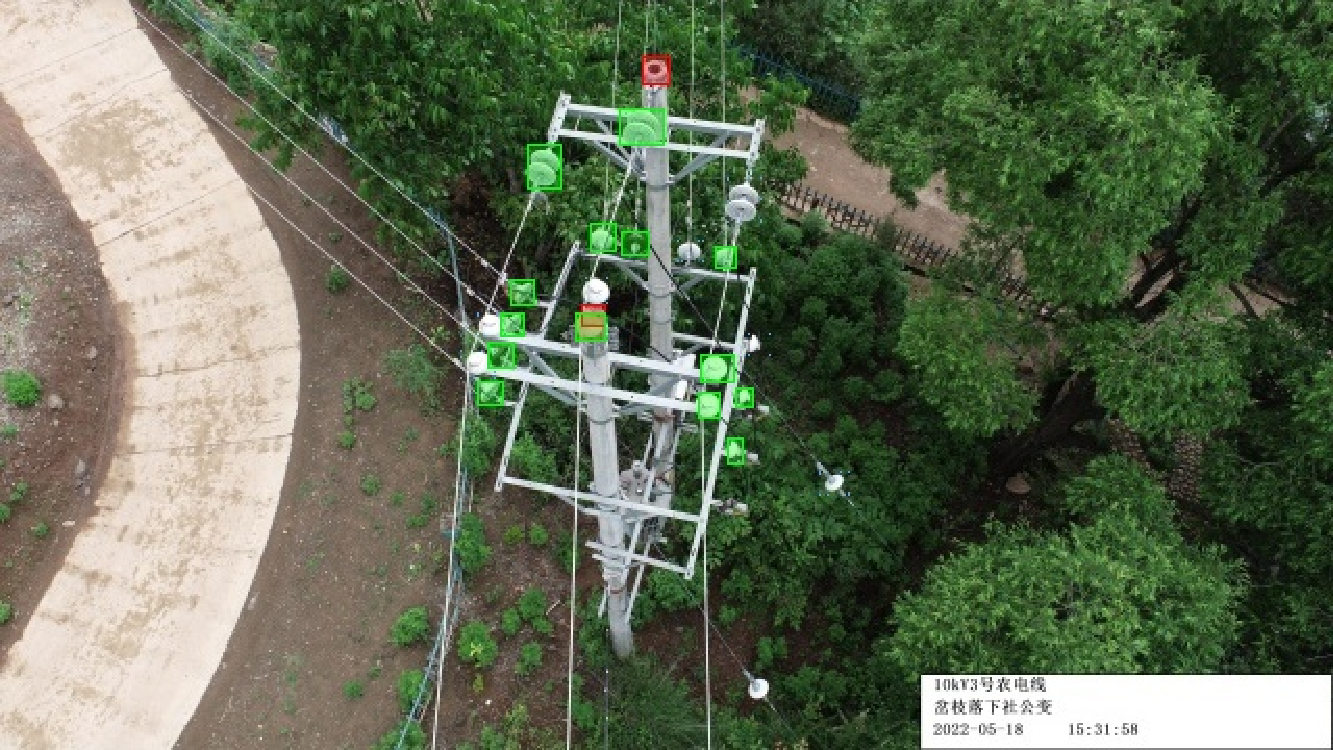}%
	}\hfill
	\subfloat[]{%
		\includegraphics[width=0.24\textwidth]{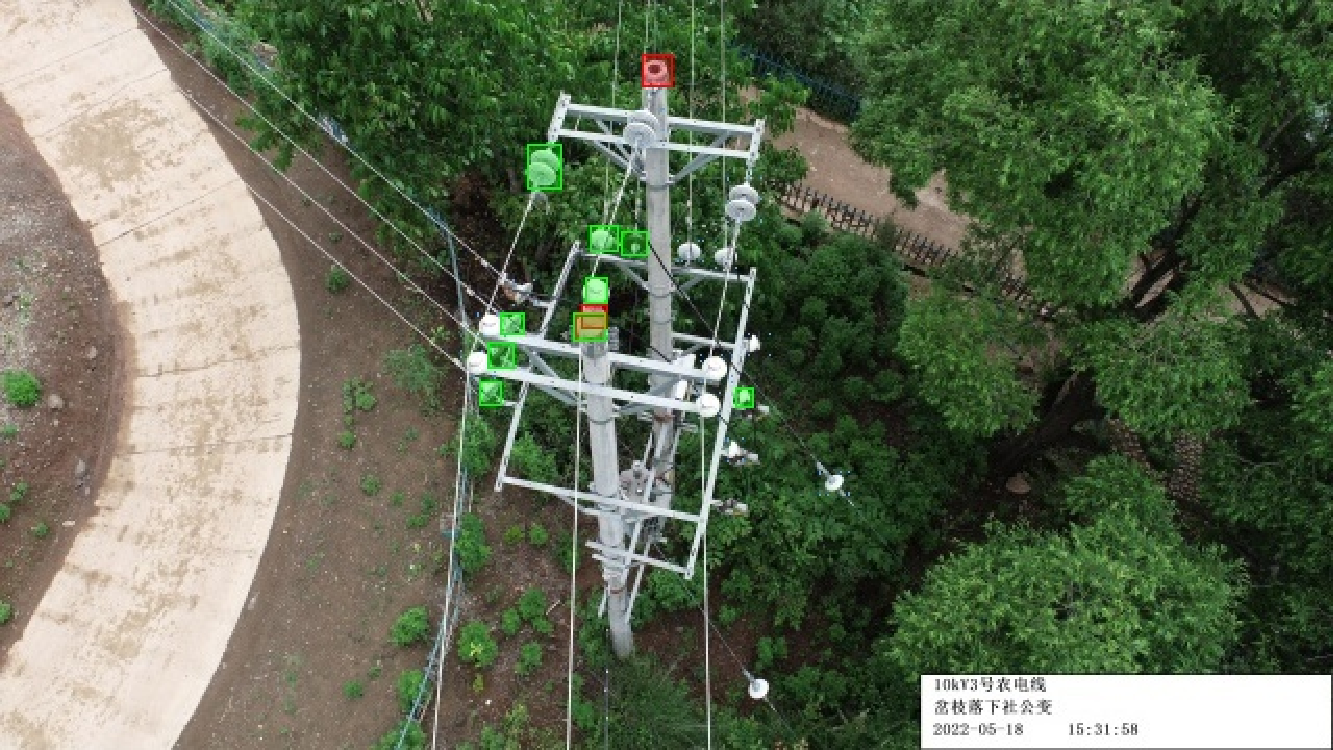}%
	}\hfill
	\subfloat[]{%
		\includegraphics[width=0.24\textwidth]{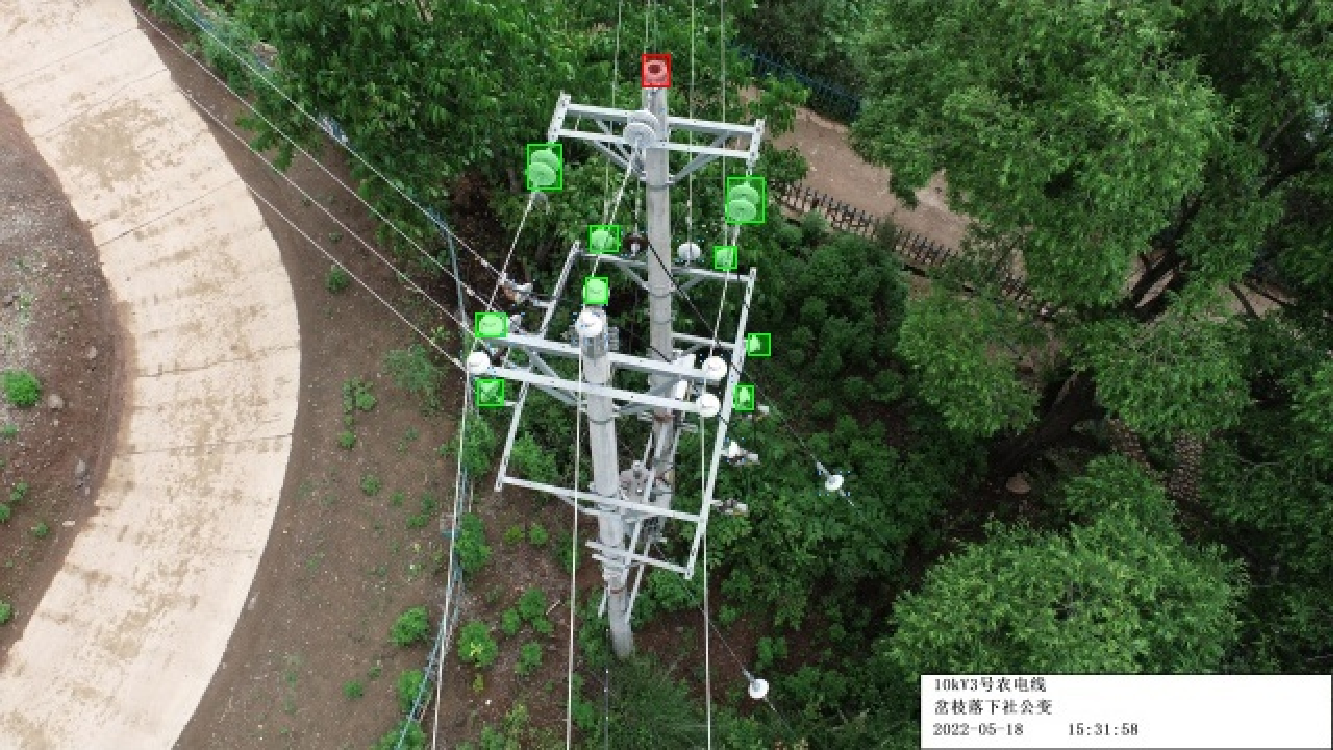}%
	}\hfill
	\subfloat[]{%
		\includegraphics[width=0.24\textwidth]{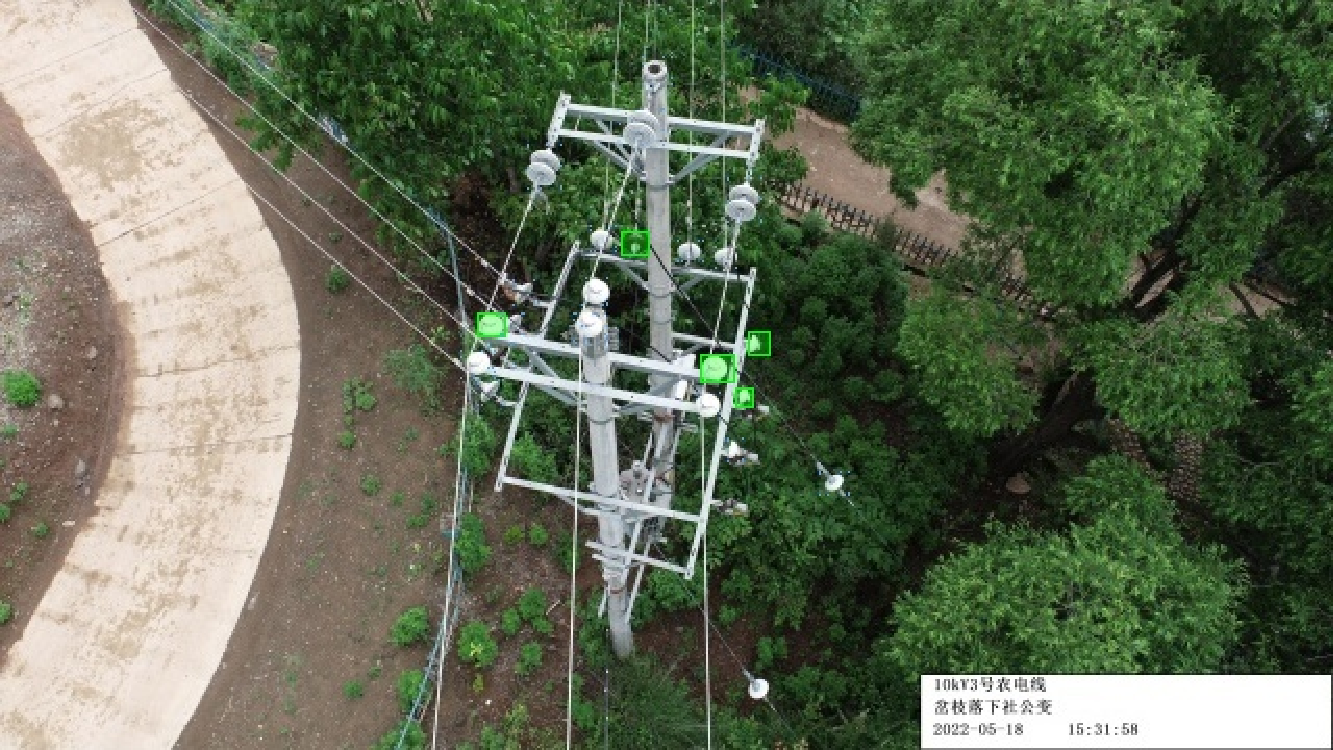}%
	}
	\caption{Qualitative comparison of detection results on four representative test samples. Each row corresponds to a different scene, and columns from left to right show predictions from CMAFNet, DINO, RT-DETR-R18, and YOLOv5s, respectively. Detection outputs are rendered as filled regions to facilitate visual discrimination under dense object distributions.}
	\label{fig:results}
\end{figure*}

Fig.~\ref{fig:results} presents qualitative comparisons on four representative test samples containing densely distributed small-scale defects. Each sample is cropped to focus on regions of high defect concentration, with detection results rendered as filled regions for visual clarity. CMAFNet consistently produces more complete and spatially coherent detections. DINO tends to generate fragmented predictions when multiple defects appear in close proximity, as evidenced by incomplete coverage in rows (b) and (c)---likely because the global attention mechanism has difficulty resolving fine-grained spatial boundaries without explicit depth guidance. RT-DETR-R18 exhibits pronounced false negatives, missing a substantial number of small targets across all scenes due to relatively coarse feature resolution at the detection head. YOLOv5s achieves reasonable coverage on larger defects but struggles with the smallest instances, producing noticeably lower recall in high-density regions. In contrast, CMAFNet maintains robust performance across varying object scales and spatial densities, yielding predictions closely aligned with the ground-truth annotations. These visual observations corroborate the quantitative findings and suggest that cross-modal fusion leverages complementary geometric cues from depth imagery to compensate for ambiguities inherent in RGB-only representations.

\begin{figure*}[h]
	\centering
	\includegraphics[width=0.95\textwidth]{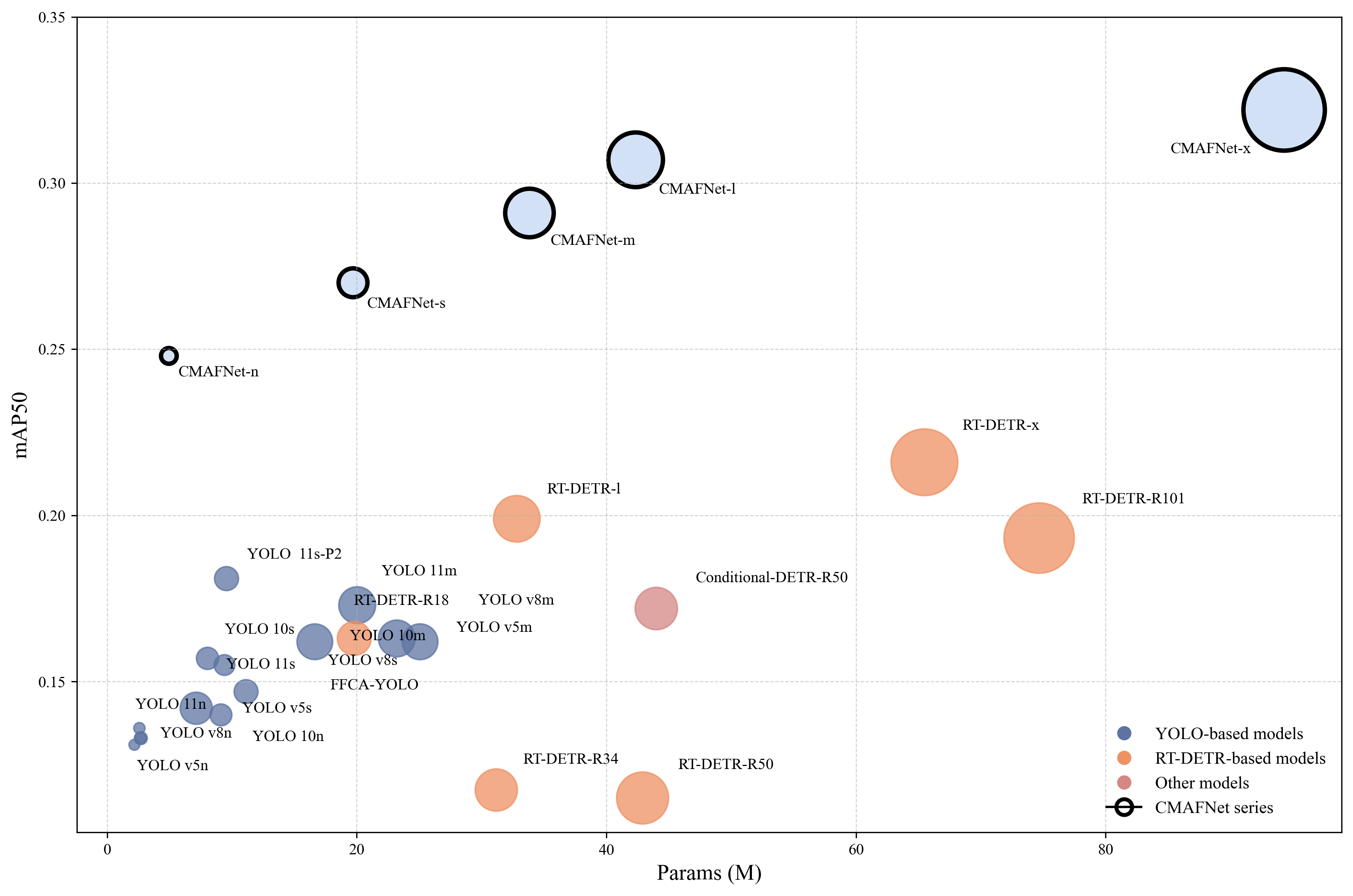}
	\caption{Accuracy-complexity trade-off visualization. The horizontal axis denotes parameters (millions), the vertical axis represents mAP$_{50}$, and bubble area is proportional to GFLOPs. CMAFNet variants are highlighted with black boundaries.}
	\label{fig:complexity_accuracy_tradeoff}
\end{figure*}

Fig.~\ref{fig:complexity_accuracy_tradeoff} visualizes the accuracy-complexity trade-off across all evaluated methods. YOLO-based methods cluster in the lower-left region with high inference speed but limited accuracy, while transformer-based methods occupy the upper-right region with higher accuracy but disproportionately large computational requirements. CMAFNet variants trace a favorable trajectory through the upper portion of the plot, achieving accuracy comparable to or exceeding that of the largest transformer models while maintaining parameter counts and GFLOPs closer to those of mid-sized YOLO detectors. From an inference perspective, CMAFNet-n achieves 227.6 FPS with an mAP$_{50}$ of 0.248, making it suitable for real-time deployment on resource-constrained UAV platforms. CMAFNet-x maintains 62.6 FPS---substantially faster than RT-DETR-x (14.9 FPS) and RT-DETR-l (38.5 FPS)---while achieving markedly higher accuracy.

\begin{figure}[h]
	\centering
	\includegraphics[width=0.85\columnwidth]{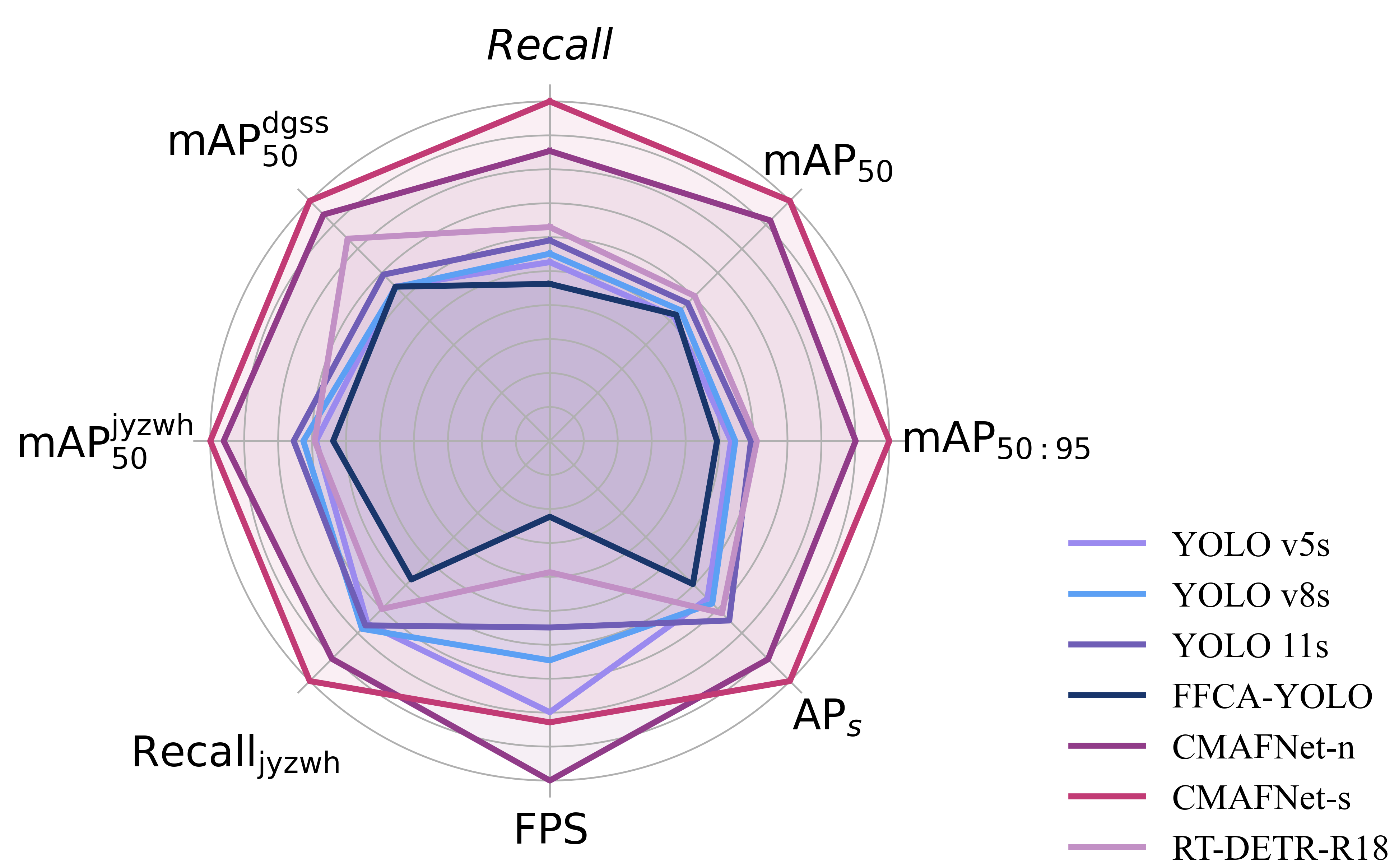}
	\caption{Multi-dimensional performance profiling via radar chart. Eight normalized indicators are visualized, with larger enclosed areas indicating more balanced detection capability.}
	\label{fig:radar_comparison}
\end{figure}

Fig.~\ref{fig:radar_comparison} presents a radar-chart visualization profiling eight normalized performance indicators: overall recall, mAP$_{50}$, mAP$_{50:95}$, AP$_s$, inference speed, and category-specific metrics for \textit{jyzwh} and \textit{dgss} defects. YOLO-based methods exhibit elongated profiles extending toward the FPS axis, reflecting optimization for speed at the expense of accuracy. Transformer-based methods display more balanced but smaller profiles, with contracted regions along the FPS axis indicating computational intensity. CMAFNet exhibits the largest enclosed area among all methods, with pronounced expansion along the recall, mAP, and category-specific axes while maintaining competitive inference speed. This comprehensive profiling confirms that CMAFNet offers superior and well-balanced performance across diverse evaluation dimensions, validating cross-modal alignment and fusion as an effective strategy for RGB-D transmission-line defect detection.

\subsection{Visualization Analysis via Feature Heatmaps}

\begin{figure*}[h]
	\centering
	\begin{subfigure}{0.32\textwidth}
		\centering
		\includegraphics[width=\linewidth]{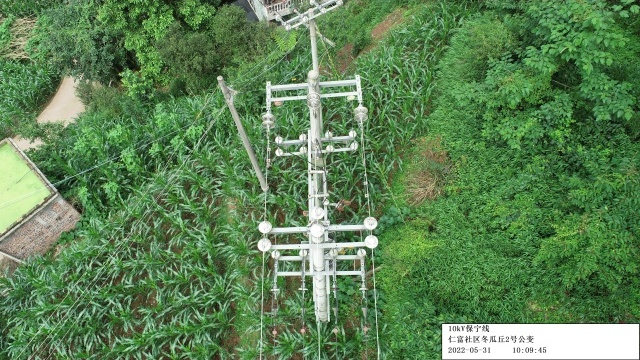}
		\caption{Original input image}
	\end{subfigure}
	\hfill
	\begin{subfigure}{0.32\textwidth}
		\centering
		\includegraphics[width=\linewidth]{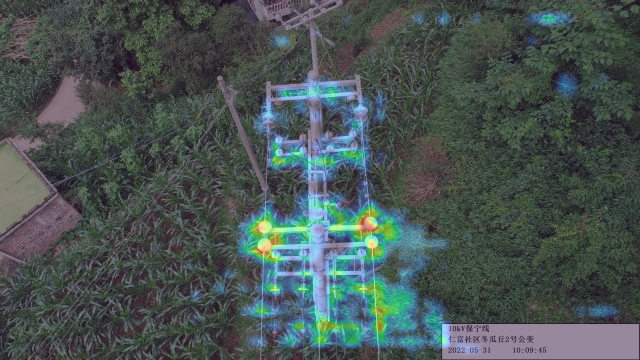}
		\caption{Baseline (w/o SRM \& CSIF)}
	\end{subfigure}
	\hfill
	\begin{subfigure}{0.32\textwidth}
		\centering
		\includegraphics[width=\linewidth]{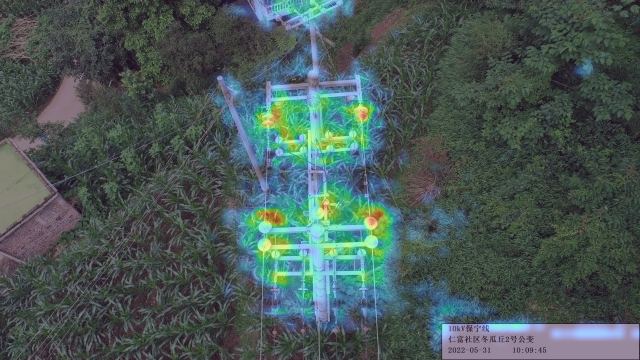}
		\caption{CMAFNet (with SRM \& CSIF)}
	\end{subfigure}
	\caption{Feature activation heatmaps extracted from the $P_3$/8-small detection layer. Warmer colors indicate higher activation intensity. The baseline model exhibits diffuse attention spread across background regions, whereas the complete CMAFNet with SRM and CSIF produces spatially concentrated activation patterns that align closely with defect locations, demonstrating that the proposed modules effectively guide the network to focus on genuine targets.}
	\label{fig:heatmap_comparison}
\end{figure*}

To provide intuitive insight into how the proposed modules influence feature learning, we visualize activation heatmaps extracted from the $P_3$/8-small detection layer, which is responsible for detecting the smallest defects in the multi-scale detection head. Fig.~\ref{fig:heatmap_comparison} presents a representative comparison between the baseline model (without SRM and CSIF) and the complete CMAFNet architecture.

The baseline model without SRM and CSIF exhibits diffuse activation patterns that spread across extensive background regions, including tower lattice structures, sky areas, and conductor segments without defects. This scattered attention distribution reflects the inability of the network to distinguish semantically relevant regions from visually similar but non-defective areas in the absence of dictionary-based purification and global context modeling mechanisms. The unfocused activations lead to two detrimental consequences: genuine defect regions receive insufficient attention weight relative to their surroundings, increasing the likelihood of missed detections, while irrelevant background regions accumulate spurious activations that contribute to false positive predictions.

In contrast, the complete CMAFNet with SRM and CSIF produces markedly different activation characteristics. The heatmap demonstrates spatially concentrated activation patterns that align closely with actual defect locations, exhibiting both precise localization and comprehensive coverage of target regions. Several distinguishing features merit attention. First, the activation intensity peaks sharply at defect centers while decaying rapidly toward the surrounding background, indicating that the network has learned to discriminate defect boundaries with high spatial precision. Second, the activation coverage extends across the full spatial extent of each defect rather than fragmenting into disconnected patches, suggesting that the global context modeling in CSIF enables the network to perceive defects as coherent semantic entities rather than isolated local patterns. Third, background suppression is substantially enhanced, with tower structures, conductors, and sky regions exhibiting minimal activation despite their visual prominence in the input image.

These visualization results corroborate the quantitative findings from the ablation study and illuminate the underlying mechanisms through which SRM and CSIF improve detection performance. The dictionary-based semantic recomposition in SRM projects features into a normalized space where modality-specific noise and incidental textures are suppressed, enabling cleaner semantic representations that activate more selectively on genuine defect patterns. Simultaneously, the global attention mechanism in CSIF establishes long-range contextual relationships that help the network distinguish defects from structurally similar but semantically distinct background elements by referencing the expected spatial arrangements of transmission-line components. The combined effect manifests as the focused, target-aligned activation patterns observed in the visualization, providing interpretable evidence that the proposed modules successfully guide the network toward more discriminative feature learning for small defect detection in complex transmission-line imagery.

\subsection{Ablation Study}

This section evaluates the contribution of each proposed module through controlled ablation experiments. We progressively remove the Semantic Recomposition Module (SRM) and the Contextual Semantic Integration Framework (CSIF) from the complete CMAFNet architecture, with results summarized in Table~\ref{tab:ablation}. To reduce training time and computational resource consumption, all ablation experiments in this section and the subsequent modality ablation study (Section~\ref{sec:modality_ablation}) are conducted on CMAFNet-n, the lightweight variant of the proposed architecture.

\begin{table*}[h]
	\centering
	\caption{Ablation study of CMAFNet under different module configurations. ``Wo'' denotes removing the corresponding module. The best performance for each metric is highlighted in bold.}
	\label{tab:ablation}
	\setlength{\tabcolsep}{6pt}
	\begin{tabular*}{\textwidth}
		{@{\hspace{6pt}\extracolsep{\fill}}lcccccccc@{\hspace{6pt}}}
		\toprule
		Configuration & $R$ & mAP$_{50}$ & mAP$_{50:95}$ & AP$_s$ & Recall$_{\mathrm{jyzwh}}$ & Recall$_{\mathrm{dgss}}$ & mAP$_{50}^{\mathrm{jyzwh}}$ & mAP$_{50}^{\mathrm{dgss}}$ \\
		\midrule
		CMAFNet & \textbf{0.240} & \textbf{0.248} & \textbf{0.109} & \textbf{0.090} & \textbf{0.450} & \textbf{0.349} & \textbf{0.438} & \textbf{0.396} \\
		Wo CSIF & 0.222 & 0.243 & 0.104 & 0.087 & 0.428 & 0.337 & 0.418 & 0.371 \\
		Only RGB SRM & 0.226 & 0.235 & 0.108 & 0.087 & 0.437 & 0.335 & 0.432 & 0.378 \\
		Only X SRM & 0.213 & 0.226 & 0.104 & 0.082 & 0.425 & 0.333 & 0.420 & 0.367 \\
		Wo All SRM & 0.224 & 0.236 & 0.105 & 0.086 & 0.433 & 0.320 & 0.426 & 0.381 \\
		Wo All SRM/CSIF & 0.217 & 0.214 & 0.097 & 0.074 & 0.374 & 0.312 & 0.359 & 0.330 \\
		\bottomrule
	\end{tabular*}
\end{table*}

The complete CMAFNet achieves the best performance across all metrics, with mAP$_{50}$ of 0.248, mAP$_{50:95}$ of 0.109, and AP$_s$ of 0.090. These results validate the design rationale behind SRM and CSIF: dictionary-based semantic purification enables the network to activate candidate regions for weak-textured and geometrically subtle defects that conventional approaches tend to miss, thereby improving recall and small-object detection in transmission-line inspection scenarios dominated by small targets.

Removing CSIF causes mAP$_{50}$ to drop from 0.248 to 0.243 and AP$_s$ from 0.090 to 0.087, while recall decreases from 0.450 to 0.428 for insulator contamination and from 0.349 to 0.337 for damaged pole. These results demonstrate the necessity of CSIF: without global context modeling at the $P_5$ fusion stage, the network loses the capacity to exploit structural priors---such as the regular arrangement of insulator strings and the spatial relationships between fittings and conductors---for semantic disambiguation of small defects embedded in complex backgrounds.

The single-branch SRM configurations reveal the complementary roles of texture and geometric information. Retaining SRM only on the RGB branch yields mAP$_{50}$ of 0.235, whereas restricting SRM to the depth branch produces inferior results with mAP$_{50}$ of 0.226. This asymmetry indicates that RGB imagery remains the dominant information source for small defect discrimination, while depth provides supplementary geometric cues including surface protrusions and boundary discontinuities. When semantic recomposition operates solely on depth features, sensor-specific artifacts---quantization noise, edge bleeding, and missing values at specular surfaces---amplify representational uncertainty, underscoring the necessity of applying SRM to both modalities for statistically aligned feature representations.

Removing all SRM instances while preserving CSIF causes mAP$_{50}$ to decline to 0.236 and the recall of damaged pole to drop from 0.349 to 0.320. This result demonstrates that global context modeling alone cannot compensate for the absence of multi-scale semantic purification. The architecture deliberately positions SRM at branch $P_3$/$P_4$ levels for intra-modal purification and at fused $P_4$/$P_5$ levels for cross-modal enhancement; disrupting this ``purify-then-fuse-then-enhance'' pipeline forces CSIF to operate on statistically heterogeneous features, limiting the effectiveness of attention mechanisms in establishing meaningful cross-modal correspondences.

The most substantial degradation occurs when both modules are removed, with mAP$_{50}$ dropping to 0.214 and AP$_s$ to 0.074---representing relative decreases of 13.7\% and 17.8\%, respectively. Notably, the ablation results reveal a synergistic effect between the two modules: CSIF alone contributes a 2.0\% improvement in mAP$_{50}$ and SRM collectively contributes 4.8\%, yet their combined integration yields a total gain of 13.7\%, demonstrating non-linear complementary benefits. This synergy arises because SRM addresses the cross-modal distribution instability that impedes semantic alignment, while CSIF resolves the limitation of purely local fusion operations that lack long-range structural constraints.

Regarding computational cost, the complete CMAFNet requires only 4.9M parameters and 12.4 GFLOPs. Since removing both modules still requires 5.0M parameters and 11.0 GFLOPs, the proposed modules achieve significant performance improvements with minimal additional overhead.

\subsection{Modality Ablation Study}
\label{sec:modality_ablation}

Table~\ref{tab:modality_ablation} presents the modality ablation results, quantifying the individual and combined contributions of RGB and depth inputs to detection performance.

The RGB-only configuration substantially outperforms the depth-only baseline across all metrics, with mAP$_{50}$ of 0.193 versus 0.151 and AP$_s$ of 0.073 versus 0.046. This disparity confirms that appearance information remains the primary discriminative source for transmission-line defect detection, as RGB imagery captures texture variations, color anomalies, and fine-grained surface patterns that depth sensors cannot reliably resolve. The depth-only setting nonetheless achieves non-trivial performance, indicating that geometric cues alone provide meaningful supervision for defect localization despite inherent sensor limitations such as quantization noise and edge artifacts.

Integrating both modalities through CMAFNet yields consistent and substantial improvements over either single-modality baseline. The fused configuration achieves mAP$_{50}$ of 0.244, representing gains of 26.4\% over the RGB-only setting and 61.6\% over the depth-only setting. More notably, AP$_s$ increases from 0.073 to 0.089, demonstrating a 21.9\% relative improvement in small-object detection---the predominant challenge in transmission-line inspection where 94.5\% of defect instances qualify as small objects. The recall metrics exhibit similar trends: insulator contamination recall rises from 0.367 (RGB) to 0.477 (RGB+Depth), and damaged pole recall improves from 0.310 to 0.365. These category-specific gains indicate that depth information provides complementary boundary and surface discontinuity cues that enhance the detection of geometrically subtle defects exhibiting low chromatic contrast against cluttered backgrounds.

The precision-recall trade-off merits attention. While the RGB-only configuration achieves the highest precision (0.677), it exhibits low recall (0.187), indicating a conservative detection strategy that misses numerous true positives. The fused model accepts a moderate precision reduction (0.452) in exchange for substantially higher recall (0.232), yielding a net improvement in overall detection performance as reflected in mAP metrics. This trade-off aligns with practical inspection requirements where false negatives carry greater risk than false positives.

From a computational perspective, the multimodal configuration introduces marginal overhead---5.5M parameters and 12.9 GFLOPs compared to 4.9M and 12.4 GFLOPs for the single-modality baselines---while maintaining real-time inference at 222.3 FPS. These results collectively validate both the effectiveness of depth as a complementary modality and the efficiency of the proposed cross-modal alignment and fusion strategy.

\begin{table*}[h]
	\centering
	\caption{Modality ablation of CMAFNet on the TL-RGBD dataset. Best results for each metric are highlighted in bold.}
	\label{tab:modality_ablation}
	\footnotesize
	\setlength{\tabcolsep}{3pt}
	\begin{tabularx}{\textwidth}{l c c *{11}{>{\centering\arraybackslash}X}}
		\toprule
		Modality & Params(M) & GFLOPs & P & R & mAP$_{50}$ & mAP$_{50:95}$ & AP$_s$ & AP$_m$ & FPS & Recall$_{jyzwh}$ & Recall$_{dgss}$ & mAP$_{50}^{jyzwh}$ & mAP$_{50}^{dgss}$ \\
		\midrule
		RGB       & \textbf{4.9} & \textbf{12.4} & \textbf{0.677} & 0.187 & 0.193 & 0.089 & 0.073 & 0.046 & \textbf{228.2} & 0.367 & 0.310 & 0.328 & 0.326 \\
		Depth     & \textbf{4.9} & \textbf{12.4} & 0.497 & 0.173 & 0.151 & 0.060 & 0.046 & 0.034 & 216.6 & 0.323 & 0.272 & 0.265 & 0.213 \\
		RGB+Depth & 5.5 & 12.9 & 0.452 & \textbf{0.232} & \textbf{0.244} & \textbf{0.103} & \textbf{0.089} & \textbf{0.071} & 222.3 & \textbf{0.477} & \textbf{0.365} & \textbf{0.431} & \textbf{0.391} \\
		\bottomrule
	\end{tabularx}
\end{table*}

\section{Conclusion}
\label{sec:conclusion}

This paper has presented CMAFNet, a cross-modal alignment and fusion network that follows a principled \emph{purify-then-fuse} paradigm for RGB-D transmission-line defect detection. The central insight motivating this design is that multimodal fusion operating on noisy, statistically misaligned representations inherits and amplifies their deficiencies rather than compensating for them---a problem that is especially acute when the vast majority of targets are small objects occupying only a few dozen pixels. CMAFNet addresses this challenge through two complementary modules. The Semantic Recomposition Module projects features through a learned bottleneck with position-wise normalization, suppressing modality-specific artifacts---illumination variation and specular noise in RGB, quantization errors and edge bleeding in depth---while narrowing the inter-modal distribution gap before integration. The Contextual Semantic Integration Framework then establishes global spatial dependencies through partial-channel attention at the deepest fusion level, enabling the network to exploit structural priors such as the regular arrangement of insulator strings and the spatial relationships between fittings and conductors for disambiguating small defects from visually similar background elements, without the detail erosion that full-channel attention would impose on fine-grained localization.

Extensive experiments on the TL-RGBD benchmark, where 94.5\% of annotated instances qualify as small objects, validate the effectiveness of the proposed approach. The full-scale variant CMAFNet-x achieves 32.2\% mAP$_{50}$ and 12.5\% AP$_s$, surpassing the strongest baseline by 9.8 and 4.0 percentage points respectively, while the lightweight configuration CMAFNet-n attains 24.8\% mAP$_{50}$ at 228 FPS with only 4.9M parameters---meeting the throughput requirements of real-time UAV-based inspection. Ablation studies reveal a synergistic interaction between SRM and CSIF: their combined integration yields a 13.7\% relative improvement in mAP$_{50}$, substantially exceeding the sum of their individual contributions, confirming that semantic purification and global context modeling address distinct yet mutually reinforcing bottlenecks. Modality ablation further demonstrates that depth provides complementary geometric cues---particularly for boundary disambiguation and low-contrast defect localization---that RGB appearance alone cannot capture, with the fused configuration achieving a 26.4\% relative gain in mAP$_{50}$ over the RGB-only baseline. Feature heatmap visualizations corroborate these quantitative findings, showing that the proposed modules transform diffuse, background-spread activations into spatially concentrated patterns tightly aligned with defect locations.

Despite these encouraging results, several limitations warrant discussion. The current architecture restricts cross-modal fusion to $P_4$ and $P_5$, deliberately excluding depth at $P_3$ to avoid injecting high-noise sensor signals into the feature map most critical for small-object localization. While this design choice proves effective under the noise characteristics of the Intel RealSense D435i sensor used in TL-RGBD, it forecloses the possibility of leveraging fine-grained geometric detail at higher resolutions---a limitation that may become significant as depth sensing hardware improves in precision and spatial fidelity. A noise-adaptive gating mechanism that dynamically modulates depth contribution at each pyramid level according to estimated signal quality would offer a more flexible alternative to the current fixed exclusion. Furthermore, the absolute detection performance, though substantially improved over prior methods, remains modest in terms of mAP$_{50:95}$, reflecting the intrinsic difficulty of achieving precise localization for targets that span fewer than $32 \times 32$ pixels. Advancing small-object AP beyond the current level will likely require joint innovation in both feature representation and label assignment strategies that account for the spatial uncertainty inherent in bounding-box supervision at this scale.

From a broader perspective, the \emph{purify-then-fuse} paradigm validated here is not restricted to the RGB-depth pairing investigated in this work. Transmission-line inspection increasingly involves heterogeneous sensor suites---including thermal infrared cameras for hotspot detection and LiDAR for three-dimensional structural modeling---whose integration faces analogous challenges of distributional mismatch and modality-specific noise. Extending the proposed framework to accommodate additional modalities, potentially through modality-agnostic purification modules that share a common bottleneck architecture while learning modality-specific projection parameters, constitutes a natural direction for future research. Equally important is the evaluation of CMAFNet on diverse infrastructure inspection benchmarks beyond transmission lines, where the combination of small-object prevalence and complex backgrounds similarly constrains single-modal approaches. Finally, while the lightweight CMAFNet-n variant already demonstrates real-time capability on GPU hardware, deploying the model on resource-constrained UAV platforms with limited power and memory budgets will require further exploration of model compression techniques---including knowledge distillation from the full-scale variant and structured pruning guided by modality-specific sensitivity analysis---to bridge the gap between laboratory throughput and field-deployable efficiency.

\bibliographystyle{elsarticle-harv}
\bibliography{ref}
\end{document}